\documentclass[11pt,a4paper]{article}
\usepackage[dvipsnames]{xcolor}
\usepackage[hyperref]{aacl-ijcnlp2020}
\usepackage{times}
\usepackage{latexsym}
\usepackage{amsmath}
\usepackage{amssymb}
\usepackage{booktabs}
\usepackage{graphicx}
\usepackage{bm}
\usepackage{comment}
\usepackage{todonotes}
\usepackage{textcomp}
\usepackage{multirow}

\usepackage{microtype}

\aclfinalcopy 
\def\aclpaperid{299} 

\newcommand{\STAB}[1]{\begin{tabular}{@{}c@{}}#1\end{tabular}}
\newcommand{\anli}{ANLI$^+$}
\newcommand{\green}[1]{\textcolor{OliveGreen}{#1}}
\newcommand{\red}[1]{\textcolor{red}{#1}}

\title{English Intermediate-Task Training Improves\\ Zero-Shot Cross-Lingual Transfer Too}

\author{Jason Phang$^{1,}$\thanks{$^*$Equal contribution.}\ \ Iacer Calixto$^{1,2,}$\footnotemark[1]\ \ ~~Phu Mon Htut$^1$\ \ ~~Yada Pruksachatkun$^1$\\ \bf Haokun Liu$^1$\ \ ~~Clara Vania$^1$\ \ Katharina Kann$^3$\ \ Samuel R. Bowman$^1$\\\\
$^1$New York University \ \ 
$^2$ILLC, University of Amsterdam\ \
$^3$University of Colorado Boulder\\
\texttt{\{jasonphang,iacer.calixto,bowman\}@nyu.edu}
}

\date{}

\begin{document}
\maketitle
\begin{abstract}
Intermediate-task training---fine-tuning a pretrained model on an \textit{intermediate} task before fine-tuning again on the target task---often improves model performance substantially on language understanding tasks in monolingual English settings.
We investigate whether English intermediate-task training is still helpful on \textit{non-}English target tasks.
Using nine intermediate language-understanding tasks,
we evaluate intermediate-task transfer in a zero-shot cross-lingual setting on the XTREME benchmark.
We see large improvements from intermediate training on the BUCC and Tatoeba sentence retrieval tasks and moderate improvements on question-answering target tasks.
MNLI, SQuAD and HellaSwag achieve the best overall results as intermediate tasks, while multi-task intermediate offers small additional improvements.
Using our best intermediate-task models for each target task, we obtain a 5.4 point improvement over \mbox{XLM-R} Large on the XTREME benchmark, setting the state of the art\footnote{The state of art on XTREME at the time of final publication in September 2020 is held by \citet{fang2020filter}, who introduce an orthogonal method.} as of June 2020.
We also investigate continuing multilingual MLM during intermediate-task training and using machine-translated intermediate-task data, but neither consistently outperforms simply performing English intermediate-task training.
\end{abstract}

\section{Introduction}


Zero-shot cross-lingual transfer involves training a model on task data in one set of languages (or language pairs, in the case of translation) and evaluating the model on the same task in unseen languages (or pairs).
In the context of natural language understanding tasks, this is generally done using a pretrained multilingual language-encoding model such as mBERT \citep{devlin2019bert}, XLM \citep{lample2019cross} or XLM-R \citep{conneau2019xlmr}
that has been pretrained with a masked language modeling (MLM) objective on large corpora of multilingual data, 
fine-tune it on task data in one language, and evaluate the tuned model on the same task in other languages.

Intermediate-task training \citep[STILTs;][]{Phang2018SentenceEO} consists of fine-tuning a pretrained model on a data-rich \textit{intermediate} task, 
before fine-tuning a second time on the target task. Despite its simplicity, this two-phase training setup has been shown to be helpful across a range of Transformer models and target tasks \citep{wang2019sesame,pruksachatkun2020intermediate}, at least within English settings.

\begin{figure*}
    \centering
    \includegraphics[width=\linewidth]{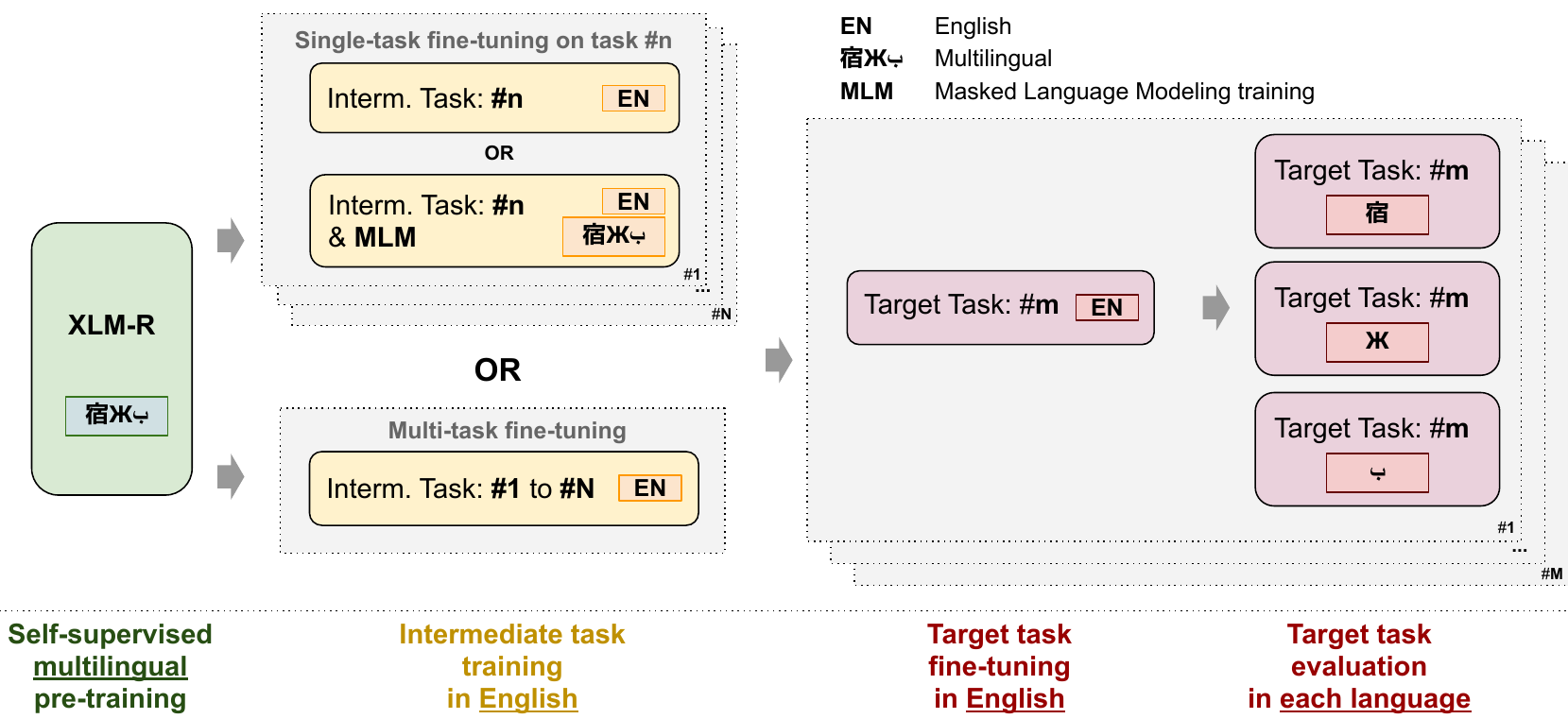}
    \caption{We investigate the benefit of injecting an additional phase of intermediate-task training on English language task data. We also consider variants using multi-task intermediate-task training, as well as continuing multilingual MLM during intermediate-task training. Best viewed in color.}
    \label{fig:crosslingual_stilts_method}
\end{figure*}

In this work, we propose to use intermediate training on English tasks to improve zero-shot cross-lingual transfer performance. Starting with a pretrained multilingual language encoder, we perform intermediate-task training on one or more English tasks, then fine-tune on the target task in English, and finally evaluate zero-shot on the same task in other languages. 

Intermediate-task training on English data introduces a potential issue: We train the pretrained multilingual model extensively on only English data before evaluating it on non-English target task data, potentially causing the model to lose the knowledge of the other languages that was acquired during pretraining \citep{Kirkpatrick2017,yogatama2019learning}. To mitigate this issue, we experiment with mixing in multilingual MLM training updates during the intermediate-task training.
In the same vein, we also conduct a case study where we machine-translate intermediate task data from English into three other languages (German, Russian and Swahili) to investigate whether intermediate training on these languages improves target task performance in the same languages.

Concretely, we use the pretrained XLM-R \citep{conneau2019xlmr} encoder and perform experiments on 9 target tasks from the recently introduced XTREME benchmark \cite{hu2020xtreme}, which aims to evaluate zero-shot cross-lingual transfer performance across diverse target tasks across up to 40 languages each.
We investigate how training on 9 different intermediate tasks, including question answering, sentence tagging, sentence completion, paraphrase detection, and natural language inference impacts zero-shot cross-lingual transfer performance.
We find the following:
\begin{itemize}
    \item Intermediate-task training on SQuAD, MNLI, and HellaSwag yields large target-task improvements of 8.2, 7.5, and 7.0 points on the development set, respectively.
    Multi-task intermediate-task training on all 9 tasks performs best, improving by 8.7 points. 
    \item Applying intermediate-task training to BUCC and Tatoeba, the two sentence retrieval target tasks that have no training data of their own, yields dramatic improvements with almost every intermediate training configuration.
    TyDiQA shows consistent improvements with many intermediate tasks, whereas XNLI does not see benefits from intermediate training.
    \item Evaluating our best performing models for each target task on the XTREME benchmark yields an average improvement of {\bf 5.4 points}, setting the state of the art as of writing.
    \item Training on English intermediate tasks
    outperforms the more complex alternatives of (i) continuing multilingual MLM during intermediate-task training, and (ii) using machine-translated intermediate-task data.
\end{itemize}

\section{Approach}

We follow a three-phase approach to training, illustrated in Figure~\ref{fig:crosslingual_stilts_method}: (i) we use a publicly available model pretrained on raw multilingual text using MLM; (ii) we perform intermediate-task training on one or more English intermediate tasks; and (iii) we fine-tune the model on English target-task training data, before evaluating it on target-task test data in each target language.

In phase (ii), our intermediate tasks have English input data.
In Section~\ref{sec:translation-approach}, we investigate an alternative where we machine-translate intermediate-task data to other languages, which we use for training.
We experiment with both single- and multi-task training for intermediate-task training.
We use target tasks from the recent XTREME benchmark for zero-shot cross-lingual transfer.

\begin{table*}[t]
  \small
  \centering
  \begin{tabular}{llrrrll}
    \toprule
    & \textbf{Name} & \textbf{$|$Train$|$}  & \textbf{$|$Dev$|$}  & \textbf{$|$Test$|$} & \textbf{Task} & \textbf{Genre/Source} \\ 
    \midrule
    \multicolumn{7}{c}{\bf Intermediate tasks}\\
    \midrule
    & \anli & 1,104,934 & 22,857 & -- & natural language inference & Misc.\\
    & MNLI & 392,702 & 20,000 & -- & natural language inference & Misc.\\
    & QQP & 363,846 & 40,430 &-- & paraphrase detection & Quora questions\\
    & SQuAD v2.0 & 130,319 & 11,873 &-- & span extraction & Wikipedia\\
    & SQuAD v1.1 & 87,599 & 10,570 & -- & span extraction & Wikipedia\\
    & HellaSwag & 39,905 & 10,042 & -- & sentence completion & Video captions \& Wikihow \\
    & CCG & 38,015 & 5,484 & -- & tagging & Wall Street Journal\\
    & Cosmos QA  & 25,588 & 3,000 & -- & question answering & Blogs \\
    & CommonsenseQA & 9,741 & 1,221 & -- & question answering & Crowdsourced responses \\     \midrule
    \multicolumn{7}{c}{\bf Target tasks (XTREME Benchmark)}\\
    \midrule
    & XNLI & 392,702 & 2,490 & 5,010 & natural language inference & Misc. \\
    & PAWS-X & 49,401 & 2,000 & 2,000 & paraphrase detection & Wiki/Quora \\
    & POS & 21,253 & 3,974 & 47--20,436 & tagging & Misc. \\
    & NER & 20,000 & 10,000 & 1,000--10,000 & named entity recognition & Wikipedia \\
    & XQuAD & 87,599 & 34,726 & 1,190 & question answering & Wikipedia\\
    & MLQA & 87,599 & 34,726 & 4,517--11,590 & question answering & Wikipedia \\
    & TyDiQA-GoldP & 3,696 & 634 & 323--2,719 & question answering & Wikipedia  \\
    & BUCC & -- & -- & 1,896--14,330 & sentence retrieval & Wiki / news \\
    & Tatoeba & -- & -- & 1,000 & sentence retrieval & Misc. \\
    \bottomrule
  \end{tabular}
  \caption{Overview of the intermediate tasks ({top}) and target tasks ({bottom}) in our experiments. For target tasks, \textit{Train} and \textit{Dev} correspond to the English training and development sets, while \textit{Test} shows the range of sizes for the target-language test sets for each task. XQuAD, TyDiQA and Tateoba do not have separate held-out development sets.
  }
  \label{tab:intermediate_target_tasks_statistics}
\end{table*}

\subsection{Intermediate Tasks}\label{sec:intermediate_tasks}

We study the effect of intermediate-task training \citep[STILTs;][]{Phang2018SentenceEO} with nine different English intermediate tasks, described in Table~\ref{tab:intermediate_target_tasks_statistics}.

We choose the tasks below based to cover a variety of task formats (classification, question answering, and multiple choice) and based on evidence of positive transfer from literature. \citet{pruksachatkun2020intermediate} shows that MNLI (of which \anli is a superset), CommonsenseQA, Cosmos QA and HellaSwag yield positive transfer to a range of downstream English-language tasks in intermediate training.
CCG involves token-wise prediction and is similar to the POS and NER target tasks. Both versions of SQuAD are widely-used question-answering tasks, while QQP is semantically similar to sentence retrieval target tasks (BUCC and Tatoeba) as well as PAWS-X, another paraphrase-detection task.

\paragraph{ANLI + MNLI + SNLI (\anli)} The Adversarial Natural Language Inference dataset \citep{nie2019adversarial} is collected using model-in-the-loop crowdsourcing as an extension of the Stanford Natural Language Inference \citep[SNLI;][]{snli:emnlp2015} and Multi-Genre Natural Language Inference \citep[MNLI;][]{mnli} corpora. We follow \citet{nie2019adversarial} and use the concatenated ANLI, MNLI and SNLI training sets, which we refer to as \anli. For all three natural language inference tasks, examples consist of premise and hypothesis sentence pairs, and the task is to classify the relationship between the premise and hypothesis as entailment, contradiction, or neutral.

\paragraph{CCG} CCGbank \citep{ccg} is a conversion of the Penn Treebank into Combinatory Categorial Grammar (CCG) derivations. The CCG supertagging task that we use consists of assigning lexical categories to individual word tokens, which together roughly determine a full parse.\footnote{If a word is tokenized into sub-word tokens, we use the representation of the first token for the tag prediction for that word as in \citet{devlin2019bert}.}

\paragraph{CommonsenseQA} CommonsenseQA \cite{commonsenseqa} is a multiple-choice QA dataset generated by crowdworkers based on clusters of concepts from ConceptNet \citep{conceptnet}.

\paragraph{Cosmos QA} Cosmos QA is multiple-choice commonsense-based \textit{reading comprehension} dataset \citep{cosmos} generated by crowdworkers, with a focus on the causes and effects of events.

\paragraph{HellaSwag} HellaSwag \citep{hellaswag} is a commonsense reasoning dataset framed as a four-way multiple choice task, where examples consist of an incomplete paragraph and four choices of spans, only one of which is a plausible continuation of the scenario. It is built using adversarial filtering \citep{zellers-etal-2018-swag, bras2020adversarial} with BERT.

\paragraph{MNLI} In additional to the full \anli, we also consider the MNLI task as a standalone intermediate task because of its already large and diverse training set.

\paragraph{QQP} Quora Question Pairs\footnote{\url{http://data.quora.com/First-Quora-DatasetRelease-Question-Pairs}\label{qqp-source}} is a paraphrase detection dataset. Examples in the dataset consist of two questions, labeled for whether they are semantically equivalent.

\paragraph{SQuAD} Stanford Question Answering Dataset \citep{rajpurkar-etal-2016-squad,rajpurkar-etal-2018-know} is a question-answering dataset consisting of passages extracted from Wikipedia articles and crowd-sourced questions and answers. In SQuAD version 1.1, each example consists of a context passage and a question, and the answer is a text span from the context. SQuAD version 2.0 includes additional questions with no answers, written adversarially by crowdworkers. We use both versions in our experiments.

\subsection{Target Tasks}
\label{sec:target_tasks}

We use the 9 target tasks from the XTREME benchmark, which span 40 different languages (hereafter referred to as the \textit{target languages}): 
Cross-lingual Question Answering \citep[\textbf{XQuAD};][]{artetxe2019cross};
Multilingual Question Answering \citep[\textbf{MLQA};][]{lewis2019mlqa};
Typologically Diverse Question Answering \citep[\textbf{TyDiQA-GoldP};][]{tydiqa};
Cross-lingual Natural Language Inference \citep[\textbf{XNLI};][]{conneau-etal-2018-xnli};
Cross-lingual Paraphrase Adversaries from Word Scrambling  \citep[\textbf{PAWS-X};][]{yang-etal-2019-paws};
Universal Dependencies v2.5 \citep{nivre:hal-01930733} {\bf POS} tagging;
Wikiann {\bf NER} \cite{pan-etal-2017-cross};
\textbf{BUCC} \cite{zweigenbaum-etal-2017-overview,zweigenbaum-etal-2018-overview}, which requires identifying parallel sentences from corpora of different languages;
and \textbf{Tatoeba} \cite{artetxe2019massively}, which involves aligning pairs of sentences with the same meaning.

Among the 9 tasks, BUCC and Tatoeba are sentence retrieval tasks that do not include training sets, and are scored based on the similarity of learned representations (see Appendix \ref{appendix:A}). XQuAD, TyDiQA and Tatoeba do not include development sets separate from the test sets.\footnote{UDPOS also does not include development sets for Kazakh, Thai, Tagalog or Yoruba.} For all XTREME tasks, we follow the training and evaluation protocol described in the benchmark paper \citep{hu2020xtreme} and their sample implementation.\footnote{\url{https://github.com/google-research/xtreme}}
Intermediate- and target-task statistics are shown in Table~\ref{tab:intermediate_target_tasks_statistics}.

\subsection{Multilingual Masked Language Modeling}
\label{sec:mlm}

Our setup requires that we train the pretrained multilingual model extensively on English data before using it on a non-English target task, which can lead to the catastrophic forgetting of other languages acquired during pretraining. We investigate whether continuing to train on the multilingual MLM pretraining objective while fine-tuning on an English intermediate task can prevent catastrophic forgetting of the target languages and improve downstream transfer performance. 

We construct a multilingual corpus across the 40 languages covered by the XTREME benchmark using Wikipedia dumps from April 14, 2020 for each language and the MLM data creation scripts from the \texttt{jiant} 1.3 library \citep{phang2020jiant}. In total, we use 2 million sentences sampled across all 40 languages using the sampling ratio from \citet{lample2019cross} with $\alpha=0.3$.

\subsection{Translated Intermediate-Task Training}
\label{sec:translation-approach}

Large-scale labeled datasets are rarely available in languages other than English for most language-understanding benchmark tasks.
Given the availability of increasingly performant machine translation models, we investigate if using machine-translated intermediate-task data can improve same-language transfer performance, compared to using English intermediate task data.

We translate training and validation data of three intermediate tasks: QQP, HellaSwag, and MNLI. We choose these tasks based on the size of the training sets and because their example-level (rather than word-level) labels can be easily mapped onto translated data. To translate QQP and HellaSwag, we use pretrained machine translation models from OPUS-MT \cite{TiedemannThottingalEAMT2020}. These models are trained with Marian-NMT \cite{mariannmt} on OPUS data \cite{TIEDEMANN12OPUS}, which integrates several resources depending on the available corpora for the language pair. For MNLI, we use the publicly available machine-translated training data of XNLI provided by the XNLI authors.\footnote{According to \citet{conneau-etal-2018-xnli}, these data are translated using a Facebook internal machine translation system.} We use German, Russian, and Swahili translations of all three datasets instead of English data for the intermediate-task training.

\section{Experiments and Results}

\subsection{Models}
\label{sec:models}

We use the pretrained XLM-R Large model \citep{conneau2019xlmr} as a starting point for all our experiments, as it currently achieves state-of-the-art performance on many zero-shot cross-lingual transfer tasks.\footnote{XLM-R Large \citep{conneau2019xlmr} is a 550m-parameter variant of the RoBERTa masked language model \citep{liu2019roberta} trained on a cleaned version of CommonCrawl on 100 languages. Notably, Yoruba is used in the POS and NER XTREME tasks but not is not in the set of 100 languages.} Details on intermediate- and target-task training can be found in Appendix \ref{appendix:A}.

\paragraph{XLM-R} For our baseline, we directly fine-tune the pretrained XLM-R model on each target task's English training data (if available) and evaluate zero-shot on non-English data, closely following the sample implementation for the XTREME benchmark.

\paragraph{XLM-R + Intermediate Task} In our main approach, as described in Figure~\ref{fig:crosslingual_stilts_method}, we include an additional intermediate-task training phase before training and evaluating on the target tasks as described above.

We also experiment with multi-task training on all available intermediate tasks.
We follow \citet{raffel2019t5paper} and sample batches of examples for each task with probability $r_m=\frac{min(e_m, K)}{\sum(min(e_m, K)}$, where $e_m$ is the number of examples in task $m$ and the constant $K=2^{17}$ limits the oversampling of data-rich tasks.

\begin{table*}[t!]
\resizebox{\textwidth}{!}{\small%
\begin{tabular}{ll ccccccccc c}
\toprule
\multicolumn{12}{c}{\textbf{Target tasks}} \\  
\midrule
&& \textbf{XNLI} & \textbf{PAWS-X} & \textbf{POS} & \textbf{NER} & \textbf{XQuAD} & \textbf{MLQA} & \textbf{TyDiQA} & \textbf{BUCC} & \textbf{Tatoeba} & \textbf{Avg.} \\
& \textbf{Metric} & \textit{acc.} & \textit{acc.} & \textit{F1} & \textit{F1} & \textit{F1 / EM} & \textit{F1 / EM} & \textit{F1 / EM} & \textit{F1} & \textit{acc.} & -- \\
& \textbf{\# langs.} & 15 & 7 & 33 & 40 & 11 & 7 & 9 & 5 & 37 & -- \\
\midrule
& \textbf{XLM-R} &   \bf 80.1   &   86.5   &   75.7   &   62.8   &   76.1 / 60.0   &   70.1 / 51.5   &   65.6 / 48.2   &   71.5   &   31.0   &   67.2  \\
\midrule \multirow{9}{*}{\STAB{\rotatebox[origin=c]{90}{\textbf{Without MLM}}}}
& \textbf{\anli} &   \red{-\ 0.8}   &   \red{-\ 0.0}   &   \red{-\ 1.4}   &   \red{-\ 3.5}   &   \red{-\ 1.1} / \red{-\ 0.5}   &   \red{-\ 0.6} / \red{-\ 0.8}   &   \red{-\ 0.6} / \red{-\ 3.0}   &   \green{+19.9}   &   \green{+48.2}   &   \green{+\ 6.6}  \\
& \textbf{MNLI} &   \red{-\ 1.2}   &   \green{+\ 1.4}   &   \underline{\red{-\ 0.7}}   &   \green{+\ 0.5}   &   \red{-\ 0.3} / \red{-\ 0.1}   &   \green{+\ 0.2} / \green{+\ 0.2}   &   \red{-\ 1.0} / \red{-\ 1.6}   &   \green{+20.0}   &   \green{+48.8}   &   \green{+\ 7.5}  \\
& \textbf{QQP} &   \red{-\ 4.4}   &   \red{-\ 4.8}   &   \red{-\ 6.5}   &   \red{-45.4}   &   \red{-\ 3.8} / \red{-\ 3.8}   &   \red{-\ 3.9} / \red{-\ 4.4}   &   \red{-11.1} / \red{-10.2}   &   \green{+17.1}   &   \green{+49.5}   &   \red{-\ 1.5}  \\
& \textbf{SQuADv1.1} &   \red{-\ 1.9}   &   \green{+\ 1.2}   &   \red{-\ 0.8}   &   \red{-\ 0.4}   &   \underline{\green{\bf +\ 1.8}} / \underline{\bf \green{+\ 2.5}}   &   \underline{\green{\bf +\ 2.2}} / \underline{\green{\bf +\ 2.6}}   &   \green{+\ 9.7} / \green{+10.8}   &   \green{+18.9}   &   \green{+41.3}   &   \green{+\ 8.1}  \\
& \textbf{SQuADv2} &   \red{-\ 1.6}   &   \underline{\green{+\ 1.9}}   &   \red{-\ 1.1}   &   \green{+\ 0.8}   &   \red{-\ 0.5} / \green{+\ 0.7}   &   \red{-\ 0.4} / \green{+\ 0.1}   &   \green{\bf +10.4} / \green{\bf +11.3}   &   \green{+19.3}   &   \green{+43.4}   &   \green{+\ 8.2}  \\
& \textbf{HellaSwag} &   \red{-\ 7.1}   &   \green{+\ 1.8}   &   \red{-\ 0.7}   &   \green{+\ 1.6}   &   \red{-\ 0.0} / \green{+\ 0.5}   &   \red{-\ 0.1} / \green{+\ 0.2}   &   \red{-\ 0.0} / \red{-\ 1.0}   &   \underline{\green{\bf +20.3}}   &   \green{+47.6}   &   \green{+\ 7.0}  \\
& \textbf{CCG} &   \red{-\ 2.6}   &   \red{-\ 3.4}   &   \red{-\ 2.0}   &   \red{-\ 1.5}   &   \red{-\ 1.5} / \red{-\ 1.3}   &   \red{-\ 1.6} / \red{-\ 1.5}   &   \red{-\ 2.8} / \red{-\ 6.2}   &   \green{+11.7}   &   \green{+41.9}   &   \green{+\ 4.1}  \\
& \textbf{CosmosQA} &   \red{-\ 2.1}   &   \red{-\ 0.3}   &   \red{-\ 1.4}   &   \red{-\ 1.5}   &   \red{-\ 0.9} / \red{-\ 1.3}   &   \red{-\ 1.5} / \red{-\ 2.0}   &   \green{+\ 0.5} / \red{-\ 0.6}   &   \green{+19.2}   &   \green{+43.9}   &   \green{+\ 6.1}  \\
& \textbf{CSQA} &   \red{-\ 2.9}   &   \red{-\ 2.8}   &   \red{-\ 1.7}   &   \red{-\ 1.6}   &   \red{-\ 1.0} / \red{-\ 1.8}   &   \red{-\ 1.0} / \red{-\ 0.6}   &   \green{+\ 3.5} / \green{+\ 2.9}   &   \green{+18.1}   &   \green{+48.6}   &   \green{+\ 6.5}  \\
& \textbf{Multi-task} &   \red{-\ 0.9}   &   \green{+\ 1.7}   &   \red{-\ 1.0}   &   \underline{\green{\bf +\ 1.8}}   &   \green{+\ 0.3} / \green{+\ 0.9}   &   \green{+\ 0.2} / \green{+\ 0.5}   &   \green{+\ 5.8} / \green{+\ 6.0}   &   \green{+19.6}   &   \underline{\green{\bf +49.9}}   &   \underline{\green{\bf +\ 8.7}}  \\
\midrule \multirow{9}{*}{\STAB{\rotatebox[origin=c]{90}{\textbf{With MLM}}}}
& \textbf{\anli} &   \red{-\ 1.1}   &   \green{+\ 1.4}   &   \underline{\green{\bf +\ 0.0}}   &   \green{+\ 0.4}   &   \red{-\ 1.9} / \red{-\ 1.7}   &   \red{-\ 0.7} / \red{-\ 0.6}   &   \green{+\ 0.9} / \green{+\ 0.5}   &   \green{+18.6}   &   \green{+46.2}   &   \green{+\ 7.1}  \\
& \textbf{MNLI} &   \red{-\ 0.7}   &   \green{+\ 1.6}   &   \red{-\ 1.6}   &   \green{+\ 1.0}   &   \red{-\ 0.7} / \green{+\ 0.1}   &   \green{+\ 0.4} / \green{+\ 0.8}   &   \red{-\ 1.8} / \red{-\ 3.2}   &   \green{+17.1}   &   \green{+44.3}   &   \green{+\ 6.6}  \\
& \textbf{QQP} &   \red{-\ 1.3}   &   \red{-\ 1.1}   &   \red{-\ 2.4}   &   \red{-\ 0.9}   &   \red{-\ 0.3} / \red{-\ 0.2}   &   \green{+\ 0.0} / \green{+\ 0.2}   &   \red{-\ 1.6} / \red{-\ 4.2}   &   \green{+14.4}   &   \green{+39.8}   &   \green{+\ 5.0}  \\
& \textbf{SQuADv1.1} &   \red{-\ 2.6}   &   \green{+\ 0.3}   &   \red{-\ 2.0}   &   \red{-\ 0.9}   &   \underline{\green{+\ 0.2}} / \underline{\green{+\ 1.6}}   &   \green{+\ 0.1} / \green{+\ 1.1}   &   \green{+\ 8.5} / \green{+\ 9.5}   &   \green{+16.0}   &   \green{+40.3}   &   \green{+\ 6.8}  \\
& \textbf{SQuADv2} &   \red{-\ 1.7}   &   \underline{\green{\bf +\ 2.1}}   &   \red{-\ 1.4}   &   \green{+\ 1.0}   &   \red{-\ 0.8} / \green{+\ 0.1}   &   \red{-\ 0.8} / \red{-\ 0.5}   &   \green{+\ 8.3} / \green{+\ 8.9}   &   \green{+15.6}   &   \green{+31.3}   &   \green{+\ 6.1}  \\
& \textbf{HellaSwag} &   \red{-\ 3.3}   &   \green{+\ 2.0}   &   \red{-\ 0.7}   &   \green{+\ 0.8}   &   \red{-\ 0.8} / \red{-\ 0.0}   &   \green{+\ 0.1} / \green{+\ 0.6}   &   \green{+\ 0.3} / \green{+\ 1.0}   &   \green{+\ 6.3}   &   \green{+22.3}   &   \green{+\ 3.1}  \\
& \textbf{CCG} &   \red{-\ 1.0}   &   \red{-\ 1.3}   &   \red{-\ 1.2}   &   \red{-\ 1.9}   &   \red{-\ 1.9} / \red{-\ 2.2}   &   \red{-\ 2.1} / \red{-\ 2.6}   &   \red{-\ 5.5} / \red{-\ 6.2}   &   \green{+\ 8.8}   &   \green{+36.1}   &   \green{+\ 3.3}  \\
& \textbf{CosmosQA} &   \red{-\ 1.0}   &   \red{-\ 1.0}   &   \red{-\ 1.6}   &   \red{-\ 3.8}   &   \red{-\ 3.1} / \red{-\ 3.3}   &   \red{-\ 3.7} / \red{-\ 4.2}   &   \red{-\ 0.6} / \red{-\ 3.2}   &   \green{+15.5}   &   \green{+42.7}   &   \green{+\ 4.7}  \\
& \textbf{CSQA} &   \underline{\red{-\ 0.5}}   &   \green{+\ 0.3}   &   \red{-\ 1.0}   &   \red{-\ 0.7}   &   \red{-\ 0.9} / \red{-\ 1.0}   &   \red{-\ 0.7} / \red{-\ 0.6}   &   \green{+\ 2.1} / \green{+\ 0.4}   &   \green{+11.6}   &   \green{+17.2}   &   \green{+\ 2.9}  \\
\midrule
\midrule
\multicolumn{12}{c}{\bf XTREME Benchmark Scores$^\dagger$}\\
\midrule
\multicolumn{2}{l}{{\bf XLM-R \citep{hu2020xtreme}}} & 79.2 &  86.4 &  72.6 &  \textbf{65.4} &  76.6 / 60.8   &  71.6 / 53.2 &  65.1 / 45.0  & 66.0 & 57.3 & 68.1\\
\multicolumn{2}{l}{{\bf XLM-R (Ours)}} &   79.5   &   86.2   &   74.0   &   62.6   &   76.1 / 60.0   &   70.2 / 51.2   &   65.6 / 48.2   &   64.5   &   31.0   &   64.8  \\
\multicolumn{2}{l}{{\bf Our Best Models}$^\ddagger$} &   {\bf 80.0}   &   {\bf 87.9}   &   {\bf 74.4}   &   64.0   &   {\bf 78.7} / {\bf 63.3}   &   {\bf 72.4} / {\bf 53.7}   &   {\bf 76.0} / {\bf 59.5}   &   {\bf 71.9}   &   {\bf 81.2}   &   {\bf 73.5}  \\
\multicolumn{2}{l}{{\bf Human \citep{hu2020xtreme}}} & 92.8 &  97.5 &  97.0 &  - &  91.2 / 82.3 &  91.2 / 82.3 &  90.1 / - &  - &  -  & - \\
\bottomrule
\end{tabular}
}
\caption{Intermediate-task training results. We compute the average target task performance across all languages, and report the median over 3 separate runs with different random seeds. Multi-task experiments use all intermediate tasks. We underline the best results per target task with and without intermediate MLM co-training, and bold-face the best overall scores for each target task. $^\dagger$: XQuAD, TyDiQA and Tatoeba do not have held-out test data and are scored using development sets in the benchmark. $^\ddagger$: Results obtained with our best-performing intermediate task configuration for each target task, selected based on the development set. The results for individual languages can be found in Appendix \ref{appendix:B}.
}
\label{tab:single_and_multi_task_results}
\end{table*}

\paragraph{XLM-R + Intermediate Task + MLM} To incorporate multilingual MLM into the intermediate-task training, we treat multilingual MLM as an additional task for intermediate training, using the same multi-task sampling strategy as above.

\paragraph{XLM-R + Translated Intermediate Task} We translate intermediate-task training and validation data for three tasks
and fine-tune XLM-R on translated intermediate-task data before we train and evaluate on the target tasks. 

\subsection{Software}
Experiments were carried out using the jiant \citep{phang2020jiant} library (2.0 alpha), based on PyTorch \citep{paszke2019pytorch} and Transformers \citep{wolf2019transformers}.

\subsection{Results}\label{sec:results}

We train three versions of each intermediate-task model with different random seeds. For each run, we compute the average target-task performance across languages, and report the median performance across the three random seeds.

\paragraph{Intermediate-Task Training}
As shown in Table~\ref{tab:single_and_multi_task_results}, no single intermediate task yields positive transfer across all target tasks. The target tasks TyDiQA, BUCC and Tatoeba see consistent gains from most or all  intermediate tasks. In particular, BUCC and Tatoeba, the two sentence retrieval tasks with no training data, benefit universally from intermediate-task training.
PAWS-X, NER, XQuAD and MLQA also exhibit gains with the additional intermediate-task training on some intermediate tasks. 
On the other hand, we find generally no or negative transfer to XNLI and POS.

\begin{table*}[t!]
\resizebox{\textwidth}{!}{
\small%
\begin{tabular}{ll ccccccccc}
\toprule
{\bf TL} & {\bf Model} & \textbf{XNLI} & \textbf{PAWS-X} & \textbf{POS} & \textbf{NER} & \textbf{XQuAD} & \textbf{MLQA} & \textbf{TyDiQA} & \textbf{BUCC} & \textbf{Tatoeba}\\
\midrule
\multirow{4}{*}{\STAB{\rotatebox[origin=c]{90}{\small{\textbf{English}}}}}
& \textbf{XLM-R} &   \bf 89.3   &   93.4   &   95.9   &   81.6   &   {\bf 86.3} / 74.2   &   81.6 / 68.6   &   70.4 / 56.6   &  --   &   --  \\
\cmidrule{2-11}
& \bf MNLI$_\texttt{en}$ &   \red{-\ 1.2}   &   \green{\bf +\ 1.6}   &   \green{+\ 0.3}   &   \green{+\ 2.6}   &   \red{-\ 2.1} / \red{-\ 1.6}   &   \green{+\ 1.1} / \green{+\ 1.4}   &   \green{+\ 1.1} / \green{+\ 1.1}   &   --   &   --  \\
& \bf QQP$_\texttt{en}$ &   \red{-\ 3.2}   &   \red{-\ 0.4}   &   \red{-\ 2.2}   &   \red{-\ 5.8}   &   \red{-\ 4.0} / \red{-\ 3.6}   &   \red{-\ 2.6} / \red{-\ 2.6}   &   \red{-\ 6.2} / \red{-\ 5.0}   &   --   &   --  \\
& \bf HellaSwag$_\texttt{en}$ &   \red{-\ 0.8}   &   \green{+\ 1.5}   &   \green{\bf +\ 0.6}   &   \green{\bf +\ 2.7}   &   \red{ -\ 0.2} / \green{\bf +\ 1.4}   &   \green{\bf +\ 1.8} / \green{\bf +\ 2.3}   &   \green{\bf +\ 1.7} / \green{\bf +\ 2.5}   &   --   &   --  \\
\midrule
\multirow{7}{*}{\STAB{\rotatebox[origin=c]{90}{\small{\textbf{German}}}}}
& \textbf{XLM-R} &   \bf 83.8   &   88.1   &   88.6   &   78.6   &   77.7 / 61.2   &  {\bf 69.1} / 52.0   &   --   &   77.7   &   63.9  \\
\cmidrule{2-11}
& \bf MNLI$_\texttt{en}$ &   \red{-\ 0.8}   &   \green{\bf +\ 0.9}   &   \red{-\ 0.1}   &   \red{-\ 0.8}   &   \red{-\ 0.3} / \red{-\ 1.0}   &   \red{-\ 1.0} / \red{-\ 0.2}   &   --   &   \green{+16.5}   &   \green{+32.7}  \\
& \bf MNLI$_\texttt{de}$ &   \red{-\ 0.4}   &   \green{+\ 0.5}   &   \red{-\ 0.3}   &   \red{-\ 0.9}   &   \green{+\ 0.2} / \red{-\ 0.3}   &   \red{-\ 2.4} / \red{-\ 2.0}   &   --   &   \green{\bf +17.0}   &   \green{+33.7}  \\
& \bf QQP$_\texttt{en}$ &   \red{-\ 2.2}   &   \red{-\ 4.2}   &   \red{-\ 3.2}   &   \red{-\ 7.3}   &   \red{-\ 4.5} / \red{-\ 4.7}   &   \red{-\ 6.7} / \red{-\ 6.4}   &   --   &   \green{+16.5}   &   \green{+32.6}  \\
& \bf QQP$_\texttt{de}$ &   \red{-\ 2.6}   &   \red{-\ 9.1}   &   \red{-\ 3.2}   &   \red{-22.9}   &   \red{-\ 6.6} / \red{-\ 5.9}   &   \red{-\ 7.7} / \red{-\ 6.6}   &   --   &   \green{+16.0}   &   \green{+33.5}  \\
& \bf HellaSwag$_\texttt{en}$ &   \red{-\ 0.3}   &   \green{+\ 0.3}   &   \green{\bf +\ 0.1}   &   \green{\bf +\ 0.5}   &   \green{\bf +\ 1.0} / \green{\bf +\ 0.2}   &   \red{ -\ 0.3} / \green{\bf +\ 0.4}   &   --   &   \green{+16.9}   &   \green{\bf +33.8}  \\
& \bf HellaSwag$_\texttt{de}$ &   \red{-\ 0.2}   &   \green{+\ 0.2}   &   \red{-\ 0.4}   &   \red{-\ 0.4}   &   \green{+\ 0.2} / \red{-\ 0.2}   &   \red{-\ 3.5} / \red{-\ 2.5}   &   --   &   \green{+16.3}   &   \green{+33.5}  \\
\midrule
\multirow{7}{*}{\STAB{\rotatebox[origin=c]{90}{\small{\textbf{Russian}}}}}
& \textbf{XLM-R} &   79.2   &   --   &   {\bf 89.5}   &   69.3   &   77.7 / 59.8   &   --   &   65.4 / 43.6   &   79.2   &   42.1  \\
\cmidrule{2-11}
& \bf MNLI$_\texttt{en}$ &   \green{\bf +\ 0.3}   &   --   &   \red{ -\ 0.0}   &   \green{+\ 0.8}   &   \green{+\ 0.1} / \green{+\ 1.5}   &   --   &   \red{-\ 1.5} / \red{-\ 4.6}   &   \green{+14.3}   &   \green{+47.1}  \\
& \bf MNLI$_\texttt{ru}$ &   \red{-\ 0.6}   &   --   &   \red{-\ 0.3}   &   \green{+\ 1.9}   &   \red{-\ 0.4} / \green{+\ 1.3}   &   --   &   \green{\bf +11.2} / \green{\bf +16.1}   &   \green{+13.1}   &   \green{+48.3}  \\
& \bf QQP$_\texttt{en}$ &   \red{-\ 0.7}   &   --   &   \red{-\ 2.9}   &   \red{-18.6}   &   \red{-\ 3.5} / \red{-\ 2.4}   &   --   &   \red{-\ 8.1} / \red{-\ 5.4}   &   \green{+14.1}   &   \green{+49.5}  \\
& \bf QQP$_\texttt{ru}$ &   \red{-\ 3.0}   &   --   &   \red{-10.6}   &   \red{-59.1}   &   \red{-\ 5.2} / \red{-\ 3.9}   &   --   &   \red{-14.4} / \red{-12.1}   &   \green{+13.3}   &   \green{+46.7}  \\
& \bf HellaSwag$_\texttt{en}$ &   \red{-\ 0.9}   &   --   &   \red{-\ 0.0}   &   \green{+\ 1.4}   &   \green{\bf +\ 0.8} / \green{\bf +\ 2.9}   &   --   &   \red{-\ 4.0} / \red{-10.6}   &   \green{\bf +14.7}   &   \green{\bf +49.9}  \\
& \bf HellaSwag$_\texttt{ru}$ &   \red{-\ 0.3}   &   --   &   \red{-\ 0.4}   &   \green{\bf +\ 2.8}   &   \green{+\ 0.2} / \green{+\ 0.2}   &   --   &   \green{+\ 8.5} / \green{+13.2}   &   \red{-71.6}   &   \red{-23.5}  \\
\midrule
\multirow{7}{*}{\STAB{\rotatebox[origin=c]{90}{\small{\textbf{Swahili}}}}}
& \textbf{XLM-R} &   {\bf 72.4}   &   --   &   --   &   69.8   &   --   &   --   &   67.2 / 48.7   &   --   &   7.9  \\
\cmidrule{2-11}
& \bf MNLI$_\texttt{en}$ &   \red{-\ 3.0}   &   --   &   --   &   \green{\bf +\ 0.6}   &   --   &   --   &   \red{-\ 0.3} / \red{-\ 0.2}   &   --   &   \green{+24.9}  \\
& \bf MNLI$_\texttt{sw}$ &   \red{-\ 1.1}   &   --   &   --   &   \red{-\ 2.4}   &   --   &   --   &   \green{+13.8} / \green{+23.4}   &   --   &   \green{\bf +47.9}  \\
& \bf QQP$_\texttt{en}$ &   \red{-\ 2.8}   &   --   &   --   &   \red{-\ 4.6}   &   --   &   --   &   \red{-12.7} / \red{-12.2}   &   --   &   \green{+27.2}  \\
& \bf QQP$_\texttt{sw}$ &   \red{-\ 7.1}   &   --   &   --   &   \red{-32.1}   &   --   &   --   &   \red{-\ 7.0} / \red{-\ 0.4}   &   --   &   \green{+41.8}  \\
& \bf HellaSwag$_\texttt{en}$ &   \red{-\ 0.4}   &   --   &   --   &   \green{+\ 0.1}   &   --   &   --   &   \red{-\ 0.9} / \red{-\ 0.4}   &   --   &   \green{+27.2}  \\
& \bf HellaSwag$_\texttt{sw}$ &   \red{-\ 9.8}   &   --   &   --   &   \green{+\ 0.4}   &   --   &   --   &   \green{\bf +15.6} / \green{\bf +26.3}   &   --   &   \red{-\ 0.5}  \\
\bottomrule
\end{tabular}
}
\caption{Experiments with translated intermediate-task training and validation data evaluated on all XTREME target tasks.
In each target language (TL) block, models are evaluated on a single target language.
We show results for models trained on original intermediate-task training data (\texttt{en}) and compare it to models trained on translated data \{\texttt{de,ru,sw}\}. 
 `--' indicates that target task data is not available for that target language.
}
\label{tab:translated_results_all_target_tasks}
\end{table*}

Among the intermediate tasks, we find that MNLI performs best; with meaningful improvements across the PAWS-X, TyDiQA, BUCC and Tatoeba tasks. \anli, SQuAD v1.1, SQuAD v2.0 and HellaSwag also show strong positive transfer performance: SQuAD v1.1 shows strong positive transfer across all three QA tasks, SQuAD v2.0 shows the most positive transfer to TyDiQA, while HellaSwag shows the most positive transfer to NER and BUCC tasks.
\anli does not show any improvement over MNLI (of which it is a superset), even on XNLI for which it offers additional directly relevant training data. This mirrors negative findings from \citet{nie2019adversarial} on NLI evaluations and \citet{bowman2020collecting} on transfer within English.
QQP significantly improves sentence retrieval-task performance, but has broadly negative transfer to the other target tasks.\footnote{For QQP, on 2 of the 3 random seeds the NER model performed extremely poorly, leading to the large negative transfer of -45.4.} CCG also has relatively poor transfer performance, consistent with \citet{pruksachatkun2020intermediate}.

Among our intermediate tasks, both SQuAD v1.1 and MNLI also serve as training sets for target tasks (for XNLI and XQuAD/MLQA respectively). While both tasks show overall positive transfer, SQuAD v1.1 actually markedly improves the performance in XQuAD and MLQA, while MNLI slightly hurts XNLI performance. We hypothesize that the somewhat surprising improvements to XQuAD and MLQA performance from SQuAD v1.1 arise due to the baseline XQuAD and MLQA models being under-trained. For all target-task fine-tuning, we follow the sample implementation for target task training in the XTREME benchmark, which trains on SQuAD for only 2 epochs. This may explain why an additional phase of SQuAD training can improve performance. Conversely, the MNLI-to-XNLI model might be over-trained, given the MNLI training set is approximately 4 times as large as the SQuAD v1.1 training set.

\paragraph{Multi-Task Training}
Multi-task training on all intermediate tasks attains the best overall average performance on the XTREME tasks, and has the most positive transfer to NER and Tatoeba tasks. However, the overall margin of improvement over the best single intermediate-task model is relatively small (only 0.3, over MNLI), while requiring significantly more training resources. Many single intermediate-task models also outperform the multi-task model in individual target tasks. \citet{wang-etal-2019-tell} also found more mixed results from a having an initial phase of multi-task training, albeit only among English language tasks across a different set of tasks. On the other hand, multi-task training precludes the need to do intermediate-task model selection, and is a useful method for incorporating multiple, diverse intermediate tasks.  

\paragraph{MLM}

Incorporating MLM during intermediate-task training shows no clear trend. It reduces negative transfer, as seen in the cases of CommonsenseQA and QQP, but it also tends to somewhat reduce positive transfer. The reductions in positive transfer are particularly significant for the BUCC and Tatoeba tasks, although the impact on TyDiQA is more mixed. On balance, we do not see that incorporating MLM improves transfer performance. 

\paragraph{XTREME Benchmark Results}

At the bottom of Table~\ref{tab:single_and_multi_task_results}, we show results obtained by \mbox{XLM-R} on the XTREME benchmark as reported by \citet{hu2020xtreme}, results obtained with our re-implementation of XLM-R (i.e. our baseline), and results obtained with our best models, which use intermediate-task configuration selected according to development set performance on each target task. Based on the results in Table~\ref{tab:single_and_multi_task_results}, which reflect the median over 3 runs, we pick the best intermediate-task configuration for each target task, and then choose the best model out of the 3 runs. Scores on the XTREME benchmark are computed based on the respective test sets where available, and based on development sets for target tasks without separate held-out test sets. We are generally able to replicate the best reported \mbox{XLM-R} baseline results, except for Tatoeba, where our implementation significantly underperforms the reported scores in \citet{hu2020xtreme}, and TyDiQA, where our implementation outperforms the reported scores. 
We also highlight that there is a large margin of difference between development and test set scores for BUCC--this is likely because BUCC is evaluated based on sentence retrieval over the given set of input sentences, and the test sets for BUCC are generally much larger than the development sets.

Our best models show gains in 8 out of the 9 XTREME tasks relative to both baseline implementations, attaining an average score of 73.5 across target tasks, a 5.4 point improvement over the previous best reported average score of 68.1. We set the state of the art on the XTREME benchmark as of June 2020, though \citet{fang2020filter} achieve higher results and hold the state of the art using an orthogonal approach at the time of our final publication in September 2020.


\paragraph{Translated Intermediate-Task Training Data}

In Table \ref{tab:translated_results_all_target_tasks}, we show results for experiments using machine-translated intermediate-training data, and evaluated on the available target-task languages. Surprisingly, even when evaluating in-language, using target-language intermediate-task data does not consistently outperform using English intermediate-task data in any of the intermediate tasks on average.

In general, cross-lingual transfer to XNLI is negative regardless of the intermediate-task or the target language.
In contrast, we observe mostly positive transfer on BUCC, and Tatoeba, with a few notable exceptions where models fail catastrophically.
TyDiQA exhibits positive transfer where the intermediate- and target-task languages aligned: intermediate training on Russian or German helps TyDiQA performance in that respective language, whereas intermediate training on English hurts non-English performance somewhat.
For the remaining tasks, there appears to be little correlation between performance and the alignment of intermediate- and target-task languages.
English language QQP already has mostly negative transfer to all target tasks except for BUCC and Tatoeba (see Table \ref{tab:single_and_multi_task_results}), and also shows a similar trend when translated into any of the three target languages.

We note that the quality of translations may affect the transfer performance. While validation performance on the translated intermediate tasks (Table~\ref{tab:translated_intermediate_performance}) for MNLI and QQP is only slightly worse than the original English versions, the performance for the Russian and Swahili HellaSwag is much worse and close to chance. Despite this, intermediate-task training on Russian and Swahili HellaSwag improve performance on PAN-X and TyDiQA, while we see generally poor transfer performance from QQP.
The interaction between translated intermediate-task data and transfer performance continues to be a complex open question.
\citet{artetxe2020translation} found that translating or back-translating training data for a task can improve zero-shot cross-lingual performance for tasks such as XNLI depending on how the multilingual datasets are created.
In contrast, we train on translated intermediate-task data and then fine-tune on a target task with English training data (excluding BUCC2018 and Tatoeba).
The authors of the XTREME benchmark have also recently released translated versions of all the XTREME task training data, which we hope will prompt further investigation into this matter.

\section{Related work}

Sequential transfer learning using pretrained Transformer-based encoders \citep{Phang2018SentenceEO} has been shown to be effective for many text classification tasks. This setup generally involves fine-tuning on a single task \citep{pruksachatkun2020intermediate,vu2020exploring} or multiple tasks \citep{liu-etal-2019-multi,wang-etal-2019-tell,raffel2019t5paper}, sometimes referred to as the intermediate task(s), before fine-tuning on the target task.
We build upon this line of work, focusing on intermediate-task training for improving cross-lingual transfer.

Early work on cross-lingual transfer mostly relies on the availability of parallel data, where one can perform translation \citep{mayhew-etal-2017-cheap} or project annotations from one language into another  \citep{hwa-etal-projection,agic-etal-2016-multilingual}. For dependency parsing, \citet{mcdonald-etal-2011-multi} use delexicalized parsers trained on source languages and labeled training data for parsing target-language data. \citet{agic-2017-cross} proposes a parser selection method to select the single best parser for a target language.

For large-scale cross-lingual transfer outside NLU, \citet{johnson-etal-2017-googles} train a single multilingual neural machine translation system with up to 7 languages and perform zero-shot translation without explicit bridging between the source and target languages. \citet{aharoni2019massively} expand this approach to cover over 100 languages in a single model. Recent works on extending pretrained Transformer-based encoders to multilingual settings show that these models are effective for cross-lingual tasks and competitive with strong monolingual models on the XNLI benchmark \citep{devlin-etal-2019-bert,lample2019cross,conneau2019xlmr,huang-etal-2019-unicoder}. More recently, \citet{artetxe2020translation} showed that cross-lingual transfer performance can be sensitive to translation artifacts arising from a multilingual datasets' creation procedure.

Finally, \citet{pfeiffer2020madx} propose adapter modules that learn language and task representations for cross-lingual transfer, which allow adaptation to languages not seen during pretraining.

\section{Conclusion}

We evaluate the impact of intermediate-task training on zero-shot cross-lingual transfer.
We investigate 9 intermediate tasks and how intermediate-task training impacts the zero-shot cross-lingual transfer to the 9 target tasks in the XTREME benchmark.

Overall, intermediate-task training significantly improves the performance on BUCC and Tatoeba, the two sentence retrieval target tasks in the XTREME benchmark, across almost every intermediate-task configuration.
Our best models obtain 5.9 and 23.9 point gains on BUCC and Tatoeba, respectively, compared to the best available XLM-R baseline scores \citep{hu2020xtreme}.
We also observed gains in question-answering tasks, particularly using SQuAD v1.1 and v2.0 as intermediate tasks, with absolute gains of 2.1 F1 for {XQuAD}, 0.8 F1 for {MLQA}, and 10.4 for F1 {TyDiQA}, again over the best available baseline scores.
We improve over XLM-R by 5.4 points on average on the XTREME benchmark.
Additionally, we found  multi-task training on all 9 intermediate tasks to slightly outperform individual intermediate training. On the other hand, we found that neither incorporating multilingual MLM into the intermediate-task training phase nor translating intermediate-task data consistently led to improved transfer performance.

While we have explored the extent to which English intermediate-task training can improve cross-lingual transfer, a clear next avenue of investigation for future work is how the choice of intermediate- and target-task languages influences transfer across different tasks.

\section*{Acknowledgments}
This project has benefited from support to SB by Eric and Wendy Schmidt (made by recommendation of the Schmidt Futures program), by Samsung Research (under the project \textit{Improving Deep Learning using Latent Structure}), by Intuit, Inc.,  by NVIDIA Corporation (with the  donation of a Titan V GPU), by Google (with the donation of Google Cloud credits). 
IC has received funding from the European Union’s Horizon 2020 research and innovation program under the Marie Sk\l{}odowska-Curie grant agreement No 838188.
This project has benefited from direct support by the NYU IT High Performance Computing Center. This material is based upon work supported by the National Science Foundation under Grant No. 1922658. Any opinions, findings, and conclusions or recommendations expressed in this material are those of the author(s) and do not necessarily reflect the views of the National Science Foundation. 

\bibliography{anthology,emnlp2020}
\bibliographystyle{acl_natbib}
\clearpage

\appendix

\section{Implementation Details}\label{appendix:A}
\subsection{Intermediate Tasks}

For intermediate-task training, we use a learning rate of 1e-5 without MLM, and 5e-6 with MLM. Hyperparameters in the Table~\ref{tab:appendix_intermediate_hyperparameters} were chosen based on  intermediate task validation performance in an preliminary search. We use a warmup of 10\% of the total number of steps, and perform early stopping based on the first 500 development set examples of each task with a patience of 30. For CCG, where tags are assigned for each word, we use the representation of first sub-word token of each word for prediction.

\begin{table}[ht!]
\centering\small
\begin{tabular}{lcc}
\toprule
Task & Batch size & \# Epochs \\
\midrule
\anli & 24 & 2 \\
MNLI & 24 & 2 \\
CCG & 24 & 15 \\
CommonsenseQA & 4 & 10 \\
Cosmos QA & 4& 15 \\
HellaSwag & 24 & 7 \\
QQP & 24 & 3 \\
SQuAD & 8 & 3 \\
MLM & 8 & -\\
Multi-task & Mixed & 3 \\
\bottomrule
\end{tabular}
\caption{Intermediate-task training configuration.}
\label{tab:appendix_intermediate_hyperparameters}
\end{table}

\subsection{XTREME Benchmark Target Tasks}

We follow the sample implementation for the XTREME benchmark unless otherwise stated. We use a learning rate of 3e-6, and use the same optimization procedure as for intermediate tasks. Hyperparameters in the Table~\ref{tab:appendix_xtreme_hyperparameters} follow the sample implementation. For POS and NER, we use the same strategy as for CCG for matching tags to tokens. For BUCC and Tatoeba, we extract the representations for each token from the 13th self-attention layer, and use the mean-pooled representation as the embedding for that example, as in the sample implementation. Similarly, we follow the sample implementation and set an optimal threshold for each language sub-task for BUCC as a similarity score cut-off for extracting parallel sentences based on the development set and applied to the test set.

We randomly initialize the corresponding output heads for each task, regardless of the similarity between intermediate and target tasks (e.g. even if both the intermediate and target tasks train on SQuAD, we randomly initialize the output head in between phases).

\begin{table}[ht!]
\centering\small
\begin{tabular}{lcc}
\toprule
Task & Batch size & \# Epochs \\
\midrule
XNLI (MNLI) & 4 & 2 \\
PAWS-X & 32 & 5 \\
XQuAD (SQuAD) & 16 & 2 \\
MLQA (SQuAD) & 16 & 2 \\
TyDiQA & 16 & 2 \\
POS & 32 & 10 \\
NER & 32 & 10 \\
BUCC & - & - \\
Tatoeba & - & - \\
\bottomrule
\end{tabular}
\caption{Target-task training configuration.}
\label{tab:appendix_xtreme_hyperparameters}
\end{table}

\section{Per-Language Results}\label{appendix:B}

\begin{table*}[t!]
\resizebox{\textwidth}{!}{\small
\begin{tabular}{llcccccccccccccccc}
\toprule
&& ar & bg & de & el & en & es & fr & hi & ru & sw & th & tr & ur & vi & zh & Avg \\
\midrule
& XLM-R & \bf 79.8 & 82.7 & \bf 83.8 & 81.3 & 89.3 & 84.4 & 83.7 & 77.3 & 79.2 & 72.4 & 77.1 & 78.9 & 72.6 & 80.0 & 79.6 & 80.1 \\
\midrule \multirow{9}{*}{\STAB{\rotatebox[origin=c]{90}{\textbf{Without MLM}}}}
& \anli & 77.5 & 82.5 & 82.3 & 80.8 & 87.6 & 83.5 & \underline{83.6} & 76.5 & 79.1 & 70.4 & 77.3 & 78.0 & 73.5 & 79.2 & 79.3 & 79.4 \\
& MNLI & 78.4 & \underline{82.8} & 83.0 & 81.3 & 88.2 & \underline{84.0} & \underline{83.6} & 77.2 & \underline{79.5} & 69.4 & 77.6 & 77.9 & 73.2 & 79.8 & 79.1 & 79.7 \\
& QQP & 77.1 & 81.0 & 81.6 & \underline{81.6} & 86.1 & 83.6 & 82.0 & 75.4 & 78.5 & 69.6 & 76.9 & 77.1 & 72.7 & 79.2 & 78.6 & 78.7 \\
& SQuAD v2.0 & 77.9 & 81.3 & 81.7 & 79.9 & 85.6 & 83.5 & 81.8 & 75.5 & 78.5 & 70.6 & 77.2 & 77.2 & \underline{73.7} & 78.9 & \underline{79.6} & 78.9 \\
& SQuAD v1.1 & 77.1 & 82.1 & 81.8 & 79.9 & 87.1 & 82.8 & 82.7 & 75.5 & 78.6 & 71.3 & 76.3 & 77.3 & 71.2 & 79.2 & 78.6 & 78.8 \\
& HellaSwag & \underline{78.6} & 82.6 & \underline{83.5} & 80.6 & \underline{88.5} & 83.7 & 83.1 & \underline{77.4} & 78.2 & \underline{72.0} & \underline{77.4} & \underline{78.7} & 73.5 & \underline{80.0} & 79.4 & \underline{79.8} \\
& CCG & 77.3 & 81.9 & 81.7 & 79.8 & 88.1 & 82.9 & 83.2 & 75.4 & 78.8 & 69.9 & 76.5 & 76.9 & 71.4 & 79.7 & 78.6 & 78.8 \\
& Cosmos QA & 77.1 & 81.1 & 81.7 & 80.1 & 87.4 & 83.2 & 81.7 & 74.3 & 77.7 & \underline{72.0} & 75.2 & 76.7 & 71.1 & 78.3 & 78.4 & 78.4 \\
& CSQA & 77.3 & 80.8 & 81.9 & 80.0 & 87.5 & 83.5 & 82.5 & 76.3 & 78.4 & 70.6 & 76.3 & 77.5 & 72.5 & 79.6 & 78.5 & 78.9 \\
& Multi-task & 76.9 & 82.2 & 82.9 & 81.0 & 88.5 & 84.4 & 82.5 & 75.8 & 79.1 & 71.1 & 77.1 & 79.1 & 72.0 & 79.6 & 79.2 & 79.4 \\
\midrule \multirow{8}{*}{\STAB{\rotatebox[origin=c]{90}{\textbf{With MLM}}}}
& \anli & 78.5 & 82.8 & \underline{\bf 83.8} & 81.5 & 89.2 & 84.1 & 82.5 & 76.5 & 79.2 & 72.7 & 77.4 & 78.6 & 72.7 & 80.7 & 80.1 & 80.0 \\
& MNLI & 78.0 & 82.9 & 83.1 & 81.1 & 88.8 & 84.3 & 83.4 & 76.7 & \underline{\bf 80.3} & 72.2 & \underline{\bf 78.4} & 79.3 & 73.4 & 80.5 & 80.2 & 80.2 \\
& QQP & 78.0 & 81.7 & 83.3 & 80.8 & 88.6 & \underline{\bf 84.5} & 82.9 & 75.9 & 78.3 & 72.2 & 77.7 & 78.6 & 72.7 & 79.9 & 78.9 & 79.6 \\
& SQuAD v2.0 & 77.5 & 82.8 & 83.3 & 80.4 & 88.8 & 83.6 & 82.7 & 76.0 & 79.6 & 71.6 & 77.0 & 78.7 & 72.9 & 79.9 & 78.9 & 79.6 \\
& SQuAD v1.1 & 77.9 & 81.7 & 82.2 & 79.7 & 87.0 & 82.8 & 82.1 & 74.4 & 78.4 & 71.2 & 76.6 & 78.1 & 71.3 & 79.0 & 78.6 & 78.7 \\
& HellaSwag & \underline{79.3} & \underline{\bf 83.5} & 83.7 & \underline{\bf 81.8} & \underline{\bf 89.6} & \underline{\bf 84.5} & \underline{\bf 84.1} & \underline{\bf 78.2} & 79.9 & 72.9 & 78.1 & \underline{\bf 80.1} & \underline{\bf 74.5} & \underline{\bf 81.3} & \underline{\bf 80.7} & \underline{\bf 80.8} \\
& CCG & 77.9 & 82.5 & 82.4 & 80.8 & 87.1 & 83.8 & 82.6 & 76.6 & 78.9 & 72.0 & 76.7 & 78.2 & 72.2 & 80.2 & 78.4 & 79.4 \\
& Cosmos QA & 78.1 & 82.7 & 82.7 & 80.4 & 87.6 & 83.9 & 82.9 & 76.2 & 79.5 & \underline{\bf 73.7} & 77.8 & 79.0 & 72.7 & 80.4 & 79.6 & 79.8 \\
& CSQA & 79.0 & 83.4 & 83.7 & 81.2 & 89.0 & 83.8 & 83.3 & 76.9 & 79.9 & 72.3 & 78.0 & 79.1 & 73.3 & 80.4 & 80.6 & 80.2 \\
\bottomrule
\end{tabular}
}
\caption{Full XNLI Results}
\label{tab:full_xnli}
\end{table*}

\begin{table*}[t!]
\centering\small
\begin{tabular}{llcccccccc}
\toprule
&& de & en & es & fr & ja & ko & zh & Avg \\
\midrule
& XLM-R & 88.1 & 93.4 & 89.2 & 89.3 & 81.8 & 81.8 & 82.0 & 86.5 \\
\midrule \multirow{9}{*}{\STAB{\rotatebox[origin=c]{90}{\textbf{Without MLM}}}}
& \anli & 88.0 & 94.1 & 89.6 & 90.7 & 82.0 & 82.2 & 81.9 & 87.0 \\
& MNLI & 89.0 & 95.0 & 90.7 & 90.9 & 82.9 & 83.8 & 84.2 & 88.1 \\
& QQP & 83.9 & 93.0 & 87.7 & 88.7 & 79.2 & 78.6 & 79.7 & 84.4 \\
& SQuADv2.0 & 88.9 & \underline{95.2} & \underline{\bf 91.7} & \underline{91.3} & 84.7 & 84.5 & \underline{\bf 85.4} & \underline{88.8} \\
& SQuADv1.1 & \underline{89.4} & 94.2 & 91.1 & 91.1 & 83.8 & 83.5 & 83.9 & 88.1 \\
& HellaSwag & 88.4 & 95.0 & 90.2 & 91.1 & \underline{84.8} & \underline{84.6} & 84.5 & 88.4 \\
& CCG & 83.5 & 92.3 & 86.5 & 88.1 & 78.0 & 77.0 & 78.6 & 83.5 \\
& Cosmos QA & 88.4 & 93.8 & 90.4 & 90.3 & 84.3 & 84.3 & 85.0 & 88.1 \\
& CSQA & 85.9 & 93.7 & 88.6 & 89.8 & 81.7 & 80.4 & 81.5 & 86.0 \\
& Multi-task & 89.0 & 95.0 & 90.2 & 91.1 & 83.8 & 83.5 & 85.5 & 88.3 \\
\midrule \multirow{8}{*}{\STAB{\rotatebox[origin=c]{90}{\textbf{With MLM}}}}
& \anli & 88.1 & 94.5 & 90.1 & 90.4 & 84.0 & 84.2 & 84.2 & 87.9 \\
& MNLI & 90.1 & \underline{\bf 95.5} & 91.3 & 91.3 & 84.4 & 84.1 & 84.5 & 88.7 \\
& QQP & 88.6 & 94.3 & 89.8 & 90.6 & 81.7 & 82.8 & 82.3 & 87.1 \\
& SQuADv2.0 & 88.9 & 95.0 & \underline{\bf 91.7} & \underline{\bf 92.0} & \underline{\bf 85.2} & 83.9 & 84.7 & 88.8 \\
& SQuADv1.1 & 89.0 & 93.8 & 90.3 & 88.9 & 82.7 & 82.2 & 82.2 & 87.0 \\
& HellaSwag & \underline{\bf 90.3} & 95.0 & 91.0 & 90.5 & 84.9 & \underline{\bf 85.9} & \underline{84.8} & \underline{\bf 88.9} \\
& CCG & 87.5 & 93.3 & 88.3 & 88.4 & 81.5 & 81.2 & 81.3 & 85.9 \\
& Cosmos QA & 88.1 & 94.0 & 89.4 & 90.0 & 82.5 & 82.4 & 82.3 & 87.0 \\
& CSQA & 88.7 & 94.1 & 89.1 & 89.8 & 82.5 & 82.9 & 82.2 & 87.0 \\
\bottomrule
\end{tabular}
\caption{Full PAWS-X Results}
\label{tab:full_pawsx}
\end{table*}

\begin{table*}[t!]
\resizebox{\textwidth}{!}{\small
\begin{tabular}{llccccccccccccccccc}
\toprule
&& af & ar & bg & de & el & en & es & et & eu & fa & fi & fr & he & hi & hu & id & it \\
\midrule
& XLM-R & 87.7 & 56.3 & 87.9 & 88.6 & 85.6 & 95.9 & 89.8 & 87.6 & 72.8 & 70.0 & 84.9 & \bf 65.5 & 68.1 & 73.2 & 81.3 & \underline{\bf 81.7} & 88.8 \\
\midrule \multirow{9}{*}{\STAB{\rotatebox[origin=c]{90}{\textbf{Without MLM}}}}
& \anli & 87.9 & 57.6 & 88.3 & 88.8 & 85.6 & 95.7 & \underline{89.4} & 87.3 & 73.4 & \underline{72.0} & 84.9 & \underline{65.4} & 70.9 & 70.1 & \underline{\bf 82.9} & 81.0 & 88.3 \\
& MNLI & 87.9 & 56.6 & 87.8 & 88.5 & 84.6 & 96.2 & 88.9 & 86.9 & 70.4 & 69.5 & 84.1 & 51.8 & 70.1 & 72.4 & 81.2 & 81.1 & 88.6 \\
& QQP & 83.9 & 52.6 & 86.0 & 85.3 & 81.7 & 93.7 & 87.7 & 82.1 & 70.1 & 66.7 & 79.3 & 62.5 & 61.1 & 62.5 & 78.3 & 79.2 & 86.8 \\
& SQuADv2.0 & 87.5 & 58.0 & 88.0 & 87.9 & 83.6 & 96.2 & 88.7 & 86.6 & 69.9 & 69.1 & 83.9 & 51.8 & \underline{\bf 71.3} & 69.7 & 82.6 & 81.0 & 89.0 \\
& SQuADv1.1 & 87.7 & \underline{58.1} & \underline{\bf 88.6} & 88.4 & 85.8 & 95.7 & \underline{89.4} & 87.2 & 73.4 & 70.1 & 84.3 & 65.1 & 70.9 & 72.2 & 81.8 & \underline{81.3} & 88.5 \\
& HellaSwag & 88.3 & 57.3 & 88.5 & 88.7 & 85.6 & \underline{\bf 96.5} & 89.2 & 87.6 & 72.6 & 69.5 & 84.7 & 52.5 & 69.6 & 74.8 & 81.6 & 81.1 & \underline{\bf 89.6} \\
& CCG & 88.2 & 56.2 & 86.5 & \underline{\bf 89.4} & \underline{85.9} & 95.8 & 87.8 & \underline{87.9} & 73.7 & 69.1 & \underline{\bf 85.6} & 53.5 & 68.8 & 75.1 & 81.8 & 80.8 & 86.8 \\
& Cosmos QA & \underline{88.4} & 56.4 & 86.2 & 88.0 & 84.4 & 95.9 & 88.9 & 87.1 & 73.5 & 71.2 & 84.5 & 65.3 & 67.5 & \underline{75.6} & 81.1 & 81.0 & 88.8 \\
& CSQA & 87.1 & 55.7 & 87.6 & 87.8 & 85.8 & 95.4 & 88.6 & 87.3 & \underline{\bf 76.4} & 69.3 & 84.7 & 64.6 & 65.3 & 67.6 & 81.2 & 80.9 & 86.6 \\
& Multi-task & 87.7 & 58.5 & 89.7 & 88.8 & 85.2 & 96.3 & 89.4 & 87.1 & 67.7 & 71.6 & 84.7 & 52.7 & 71.0 & 68.2 & 81.5 & 80.7 & 89.8 \\
\midrule
\midrule \multirow{8}{*}{\STAB{\rotatebox[origin=c]{90}{\textbf{With MLM}}}}
& \anli & 87.9 & \underline{\bf 58.4} & 88.3 & 88.9 & \underline{\bf 86.3} & 95.8 & \underline{\bf 90.3} & 87.8 & \underline{\bf 76.4} & \underline{\bf 72.5} & \underline{85.1} & 53.3 & 69.0 & 72.5 & \underline{82.4} & 80.7 & 88.6 \\
& MNLI & \underline{\bf 89.1} & 57.2 & 87.6 & 88.6 & 85.1 & 96.2 & 88.8 & 88.0 & 73.4 & 69.5 & \underline{85.1} & 52.7 & 68.0 & 76.9 & 80.6 & 80.4 & 88.7 \\
& QQP & 87.7 & 56.3 & 87.6 & 88.6 & 84.2 & 95.9 & 89.6 & \underline{\bf 88.1} & 76.3 & 71.2 & 84.5 & 59.7 & 67.5 & \underline{\bf 78.0} & 81.8 & 81.2 & 88.8 \\
& SQuADv2.0 & 88.5 & 57.8 & 87.8 & 88.5 & 85.8 & 96.2 & 89.0 & 86.1 & 74.7 & 71.0 & 84.6 & 49.1 & 68.2 & 73.2 & 81.4 & 80.8 & 85.8 \\
& SQuADv1.1 & 88.0 & 55.1 & \underline{\bf 88.6} & 88.9 & 85.3 & 95.7 & 89.7 & 85.7 & 73.5 & 70.2 & 83.5 & 64.5 & 66.7 & 74.4 & 79.7 & \underline{81.5} & 86.8 \\
& HellaSwag & 88.3 & 58.0 & 87.8 & 88.3 & 85.7 & \underline{96.4} & 87.2 & 86.8 & 74.0 & 70.2 & 84.3 & 51.5 & \underline{70.9} & 74.8 & 79.9 & 81.0 & 88.4 \\
& CCG & 88.1 & 54.5 & 86.7 & \underline{89.2} & \underline{\bf 86.3} & 95.9 & 87.5 & 87.6 & 77.2 & 71.4 & 84.0 & 64.4 & 66.3 & 76.7 & 81.1 & 81.4 & \underline{89.0} \\
& Cosmos QA & 87.5 & 57.8 & 87.7 & 88.6 & 85.5 & 95.8 & 89.5 & \underline{\bf 88.1} & 71.7 & 70.1 & 84.9 & 64.4 & 68.9 & 76.6 & 81.0 & 80.0 & 88.3 \\
& CSQA & 87.6 & 55.9 & 87.4 & 88.7 & 85.1 & 95.6 & 88.5 & 87.2 & \underline{\bf 76.4} & 70.4 & 84.2 & \underline{65.1} & 68.2 & 68.3 & 81.6 & 81.2 & 88.4 \\
\midrule
\midrule
&& ja & kk & ko & mr & nl & pt & ru & ta & te & th & tl & tr & ur & vi & yo & zh & Avg \\
\midrule
& XLM-R & 31.9 & - & 50.4 & 80.0 & 90.1 & \bf 90.2 & 89.5 & 67.1 & \bf 90.0 & - & - & 76.0 & 65.6 & 56.4 & - & 40.9 & 75.7 \\
\midrule \multirow{9}{*}{\STAB{\rotatebox[origin=c]{90}{\textbf{Without MLM}}}}
& \anli & 19.4 & - & 50.7 & 79.6 & 90.1 & 89.7 & \underline{\bf 90.0} & 69.2 & 86.6 & - & - & 75.0 & 66.2 & 55.3 & - & 27.2 & 74.8 \\
& MNLI & 38.1 & - & 50.7 & 79.1 & 90.4 & 89.7 & 89.4 & 69.4 & 86.7 & - & - & 74.8 & 67.6 & 54.4 & - & \underline{\bf 48.6} & \underline{75.4} \\
& QQP & 6.2 & - & 45.9 & 73.5 & 88.4 & 88.2 & 86.6 & 65.1 & 81.7 & - & - & 71.5 & 59.1 & 54.5 & - & 12.0 & 70.1 \\
& SQuADv2.0 & \underline{\bf 39.4} & - & \underline{50.8} & 80.5 & 90.3 & \underline{90.1} & 89.1 & 68.5 & 86.1 & - & - & 74.1 & 60.6 & 54.1 & - & 45.3 & 75.0 \\
& SQuADv1.1 & 30.9 & - & 49.7 & 78.7 & \underline{\bf 90.5} & 89.7 & 89.3 & 66.8 & 84.9 & - & - & 74.4 & 65.4 & 56.2 & - & 37.7 & 75.3 \\
& HellaSwag & 31.1 & - & 50.5 & 83.7 & 90.1 & 89.8 & 89.5 & \underline{\bf 69.7} & 86.2 & - & - & 74.2 & 67.4 & 54.5 & - & 35.1 & 75.2 \\
& CCG & 17.8 & - & 50.3 & 81.0 & 90.1 & 88.0 & 88.9 & 66.8 & 88.4 & - & - & 75.9 & 70.7 & 55.5 & - & 23.1 & 74.1 \\
& Cosmos QA & 16.4 & - & 50.3 & 77.7 & 89.9 & 89.7 & 89.4 & 67.9 & 88.1 & - & - & \underline{\bf 76.5} & \underline{69.2} & \underline{56.3} & - & 23.2 & 74.4 \\
& CSQA & 32.4 & - & 49.3 & \underline{82.8} & 89.4 & 88.5 & 88.5 & 66.9 & 86.3 & - & - & 74.5 & 63.5 & 56.0 & - & 29.6 & 74.5 \\
& Multi-task & 36.4 & - & 50.7 & 79.6 & 90.0 & 89.8 & 88.9 & 68.4 & 86.2 & - & - & 74.4 & 62.2 & 55.5 & - & 44.3 & 75.1 \\
\midrule
\midrule \multirow{8}{*}{\STAB{\rotatebox[origin=c]{90}{\textbf{With MLM}}}}
& \anli & \underline{39.0} & - & 51.2 & 80.7 & 90.2 & \underline{90.0} & \underline{89.8} & 68.7 & 87.6 & - & - & \underline{76.4} & 66.2 & 56.7 & - & \underline{45.7} & \underline{\bf 76.1} \\
& MNLI & 30.1 & - & 51.0 & 80.1 & 90.0 & 88.8 & 89.1 & \underline{68.8} & 85.5 & - & - & 75.1 & 69.6 & 55.4 & - & 38.4 & 75.1 \\
& QQP & 27.6 & - & 50.8 & 81.0 & 90.1 & 89.5 & 89.4 & 67.2 & \underline{88.0} & - & - & 76.2 & \underline{\bf 70.3} & 56.5 & - & 34.0 & 75.4 \\
& SQuADv2.0 & 35.3 & - & 51.0 & 80.2 & 89.9 & 88.1 & 89.3 & 67.1 & 84.3 & - & - & 75.5 & 68.8 & 56.9 & - & 39.0 & 75.0 \\
& SQuADv1.1 & 16.3 & - & 49.7 & 79.4 & 90.2 & \underline{90.0} & 89.2 & 68.0 & 83.3 & - & - & 75.8 & 64.6 & \underline{\bf 57.3} & - & 19.0 & 73.8 \\
& HellaSwag & 35.4 & - & 50.9 & 78.4 & 90.0 & 87.9 & 89.3 & 68.7 & 86.4 & - & - & 75.4 & 69.3 & 54.8 & - & 43.6 & 75.3 \\
& CCG & 25.7 & - & 50.7 & \underline{\bf 86.1} & 89.8 & 88.8 & 88.4 & 68.0 & 86.6 & - & - & 76.2 & 68.2 & 55.5 & - & 23.9 & 75.0 \\
& Cosmos QA & 16.5 & - & 51.0 & 80.9 & 89.7 & 88.9 & 89.0 & 67.4 & 87.9 & - & - & 76.3 & 70.1 & 56.0 & - & 19.6 & 74.5 \\
& CSQA & 30.8 & - & \underline{\bf 51.8} & 80.5 & \underline{\bf 90.5} & 89.6 & 89.0 & 66.8 & 86.5 & - & - & 74.8 & 61.9 & 56.3 & - & 31.3 & 74.8 \\
\bottomrule
\end{tabular}
}
\caption{Full POS Results. \texttt{kk}, \texttt{th}, \texttt{tl} and \texttt{yo} do not have development set data.}
\label{tab:full_udpos}
\end{table*}

\begin{table*}[t!]
\resizebox{\textwidth}{!}{\small
\begin{tabular}{llccccccccccccccccccccc}
\toprule
&& af & ar & bg & bn & de & el & en & es & et & eu & fa & fi & fr & he & hi & hu & id & it & ja & jv & ka \\
\midrule
& XLM-R & 77.7 & 47.1 & 81.9 & 74.9 & 78.6 & 76.3 & 81.6 & 74.7 & 77.2 & 61.2 & 58.2 & 78.3 & 78.3 & 50.2 & 68.7 & 80.6 & 53.7 & 80.8 & 15.6 & 56.2 & 61.4 \\
\midrule \multirow{9}{*}{\STAB{\rotatebox[origin=c]{90}{\textbf{Without MLM}}}}
& \anli & 75.4 & 52.7 & 78.1 & 72.7 & 76.4 & 76.3 & 80.9 & 71.6 & 72.8 & 52.2 & 60.7 & 75.8 & 77.4 & 49.1 & 69.6 & 79.6 & 52.7 & 78.9 & 13.1 & 54.3 & 62.1 \\
& MNLI & 76.9 & 48.3 & 80.5 & 72.8 & 77.7 & 77.9 & 84.2 & 76.9 & 78.5 & 62.1 & 58.3 & 78.7 & 81.1 & 55.1 & 69.0 & 81.1 & 55.7 & 80.8 & 16.4 & 54.2 & 68.1 \\
& QQP & 73.8 & 40.9 & 75.5 & 66.0 & 71.3 & 71.6 & 75.8 & 65.5 & 69.3 & 55.5 & 49.9 & 73.1 & 72.8 & 42.6 & 59.8 & 74.3 & 49.2 & 75.9 & 5.7 & 54.4 & 51.1 \\
& SQuADv2.0 & 76.0 & 48.0 & 81.1 & 71.8 & 78.4 & 78.2 & 84.3 & 74.7 & 78.4 & 53.9 & 56.9 & 78.9 & \bf 82.5 & 56.0 & 68.9 & 79.8 & 56.4 & 80.8 & 18.1 & 61.8 & 67.3 \\
& SQuADv1.1 & \bf 79.1 & 52.6 & 80.1 & 75.5 & 77.8 & 78.1 & 80.8 & 75.3 & 76.7 & 54.3 & 61.9 & 78.7 & 78.4 & 52.8 & 65.6 & 80.3 & 54.6 & 80.8 & 18.7 & 52.1 & 62.4 \\
& HellaSwag & 77.0 & 54.9 & 82.7 & \bf 76.6 & 79.1 & \bf 78.9 & 84.3 & \bf 77.8 & 78.0 & 58.8 & 65.0 & 77.5 & 80.3 & 57.0 & \bf 71.2 & 81.8 & 54.3 & \bf 81.4 & 19.6 & 56.9 & 70.6 \\
& CCG & 77.4 & 51.5 & 78.7 & 72.5 & 78.4 & 76.2 & 80.8 & 73.0 & 78.0 & 56.9 & 62.1 & 78.2 & 77.3 & 48.6 & 67.3 & 79.7 & 54.9 & 79.9 & 15.9 & 60.3 & 58.9 \\
& Cosmos QA & 76.6 & 49.3 & 79.2 & 76.0 & 77.8 & 76.1 & 81.2 & 73.2 & 76.6 & 59.8 & 55.8 & 77.8 & 77.0 & 46.8 & 67.8 & 79.4 & 53.2 & 80.0 & 14.1 & 55.5 & 57.8 \\
& CSQA & 77.6 & 46.1 & 78.9 & 75.4 & 78.4 & 76.2 & 81.3 & 77.3 & 75.2 & 59.8 & 61.9 & 78.0 & 78.2 & 48.9 & 67.6 & 79.6 & 55.6 & 80.1 & 11.6 & 53.8 & 57.7 \\
& Multi-task & 78.5 & 49.2 & 82.0 & 73.3 & 78.9 & 80.1 & 84.5 & 76.6 & 78.5 & 59.4 & 49.4 & 79.1 & 81.2 & 56.4 & 70.6 & 81.0 & 57.0 & 80.7 & 20.7 & 64.7 & 68.6 \\
\midrule
\midrule \multirow{8}{*}{\STAB{\rotatebox[origin=c]{90}{\textbf{With MLM}}}}
& \anli & 76.4 & 51.5 & 80.7 & 73.3 & 79.2 & 77.8 & 84.3 & 75.4 & 78.0 & 57.7 & 49.7 & 77.6 & 80.1 & 54.8 & 68.9 & 80.8 & 54.8 & 80.5 & 14.4 & 54.9 & 64.5 \\
& MNLI & 78.0 & 52.3 & 81.7 & 73.0 & 79.6 & 78.1 & 84.4 & 77.2 & 79.4 & 59.6 & 60.6 & 79.2 & 81.4 & 55.1 & 68.6 & 81.0 & 51.3 & 81.0 & 14.0 & 62.0 & 64.3 \\
& QQP & 77.1 & 46.7 & 79.0 & 72.9 & 79.4 & 76.3 & 81.9 & 74.2 & 78.7 & 61.8 & \bf 66.0 & 78.3 & 78.0 & 50.4 & 69.1 & 81.6 & 53.2 & 80.1 & 15.1 & \bf 62.6 & 60.7 \\
& SQuADv2.0 & 78.0 & 46.5 & \bf 82.8 & 71.7 & 79.0 & 77.3 & 84.2 & 74.8 & 79.0 & 61.6 & 63.3 & 79.5 & 80.0 & \bf 57.6 & 67.5 & \bf 81.9 & \bf 62.0 & 80.7 & \bf 20.0 & 62.3 & 68.2 \\
& SQuADv1.1 & 77.7 & \bf 58.0 & 81.4 & 75.2 & 78.0 & 77.4 & 82.1 & 69.6 & 76.1 & 54.1 & 58.4 & 77.5 & 78.7 & 54.8 & 67.5 & 78.8 & 49.9 & 79.5 & 14.5 & 55.9 & \bf 68.3 \\
& HellaSwag & 78.7 & 47.0 & 81.8 & 73.8 & \bf 79.7 & 78.2 & \bf 84.8 & 73.6 & 79.2 & 55.8 & 55.6 & 78.2 & 79.4 & 55.0 & 69.8 & 81.3 & 54.1 & 81.3 & 18.5 & 58.1 & 67.5 \\
& CCG & 74.5 & 46.4 & 76.7 & 74.5 & 76.9 & 75.7 & 80.5 & 72.6 & 77.7 & 58.9 & 59.6 & 77.7 & 77.0 & 48.1 & 66.3 & 80.1 & 53.4 & 78.7 & 13.8 & 57.1 & 58.2 \\
& Cosmos QA & 78.2 & 39.1 & 80.0 & 73.8 & 79.0 & 77.2 & 81.4 & 70.3 & 78.8 & 65.4 & 48.9 & 78.7 & 77.7 & 48.3 & 68.0 & 80.8 & 55.1 & 81.2 & 13.2 & 58.9 & 59.0 \\
& CSQA & 77.4 & 48.8 & 78.9 & 73.9 & 78.8 & 76.3 & 81.9 & 75.2 & \bf 79.5 & \bf 66.7 & 58.6 & \bf 79.6 & 78.5 & 47.7 & 68.2 & 81.0 & 55.3 & 81.3 & 12.2 & 60.4 & 58.9 \\
\midrule
\midrule
&& kk & ko & ml & mr & ms & my & nl & pt & ru & sw & ta & te & th & tl & tr & ur & vi & yo & zh & Avg & \\
\midrule
& XLM-R & 48.7 & 54.5 & 58.8 & 61.8 & 54.1 & 53.7 & 83.2 & 80.7 & 69.3 & 69.8 & 58.2 & 50.8 & 2.2 & 73.2 & 81.1 & 67.0 & 74.9 & 33.2 & 23.6 & 62.8 & - \\
\midrule \multirow{9}{*}{\STAB{\rotatebox[origin=c]{90}{\textbf{Without MLM}}}}
& \anli & 50.2 & 52.6 & 61.2 & 63.0 & 66.8 & 46.5 & 81.8 & 78.7 & 67.0 & 66.9 & 55.0 & 52.1 & 2.5 & 71.2 & 78.0 & 67.3 & 73.9 & 43.3 & 18.9 & 62.0 & - \\
& MNLI & 51.7 & 58.8 & 64.8 & 61.3 & 69.8 & 54.9 & 83.0 & 80.8 & 70.2 & 70.3 & 59.3 & \bf 55.4 & 1.0 & 74.8 & 80.5 & 56.9 & 78.1 & 38.9 & 25.2 & 64.2 & - \\
& QQP & 50.4 & 40.1 & 51.2 & 51.4 & 61.4 & 32.5 & 78.2 & 73.0 & 50.8 & 65.1 & 47.3 & 41.4 & 1.6 & 67.4 & 72.3 & 57.2 & 67.9 & 43.9 & 8.6 & 55.9 & - \\
& SQuADv2.0 & 49.9 & 58.1 & 61.6 & 62.5 & 72.1 & 50.0 & 83.1 & 82.3 & 70.8 & 65.4 & 62.6 & 53.6 & 0.6 & 74.8 & 80.0 & 63.2 & \bf 78.9 & 41.2 & 22.5 & 64.1 & - \\
& SQuADv1.1 & 51.8 & 57.1 & 61.7 & 59.8 & 50.4 & 52.2 & 83.3 & 80.8 & 69.8 & 69.2 & 58.3 & 49.5 & 0.8 & 71.6 & 79.1 & 58.6 & 76.3 & \bf 47.5 & 26.2 & 63.0 & - \\
& HellaSwag & 50.5 & 58.4 & 56.6 & \bf 66.6 & 72.8 & \bf 59.4 & 83.2 & \bf 82.5 & 70.8 & 69.9 & \bf 63.7 & 53.0 & 1.1 & 75.1 & 78.0 & 70.0 & 75.0 & 42.1 & \bf 29.7 & \bf 65.5 & - \\
& CCG & 52.4 & 52.7 & 57.7 & 59.6 & 52.3 & 50.0 & 82.5 & 79.0 & 67.1 & 67.0 & 55.3 & 49.1 & \bf 2.6 & 70.0 & 81.0 & 65.3 & 74.2 & 37.6 & 23.3 & 62.1 & - \\
& Cosmos QA & 48.4 & 52.4 & 60.3 & 62.1 & 56.9 & 50.2 & 82.8 & 79.5 & 67.4 & 67.8 & 57.2 & 51.4 & 1.3 & 74.6 & 80.7 & 60.8 & 74.9 & 34.8 & 19.5 & 61.8 & - \\
& CSQA & 49.7 & 52.0 & 59.1 & 62.9 & 62.4 & 46.1 & 82.5 & 80.3 & 65.4 & 69.0 & 57.1 & 51.2 & 1.8 & 73.1 & 80.2 & \bf 73.3 & 73.5 & 35.3 & 19.3 & 62.3 & - \\
& Multi-task & 53.2 & 57.8 & 60.8 & 61.0 & 69.3 & 54.2 & 83.8 & 80.8 & 69.4 & 70.6 & 58.9 & 53.7 & 2.2 & 75.2 & 77.2 & 57.7 & 75.6 & 46.1 & 30.4 & 64.7 & - \\
\midrule
\midrule \multirow{8}{*}{\STAB{\rotatebox[origin=c]{90}{\textbf{With MLM}}}}
& \anli & 52.9 & 56.8 & 60.0 & 61.1 & \bf 75.4 & 49.5 & \bf 83.4 & 80.9 & 68.3 & 71.0 & 57.2 & 49.8 & 0.9 & 74.5 & 79.0 & 59.8 & 76.3 & 31.7 & 22.5 & 63.2 & - \\
& MNLI & \bf 54.7 & 57.5 & 63.5 & 63.3 & 66.3 & 49.6 & \bf 83.4 & 81.1 & 70.3 & 72.2 & 57.0 & 53.5 & 1.1 & 74.1 & 80.9 & 61.1 & 75.1 & 43.4 & 22.8 & 64.3 & - \\
& QQP & 49.9 & 54.5 & 63.3 & 64.6 & 54.7 & 49.0 & 82.9 & 78.9 & 68.7 & 70.9 & 58.0 & 50.7 & 1.1 & 74.0 & \bf 82.3 & 70.2 & 77.1 & 40.3 & 24.9 & 63.5 & - \\
& SQuADv2.0 & 52.1 & \bf 60.8 & \bf 65.1 & 63.2 & 54.7 & 54.8 & \bf 83.4 & 80.9 & \bf 71.6 & \bf 72.6 & 63.0 & 54.1 & 0.4 & \bf 75.3 & 80.4 & 59.8 & 77.6 & 33.6 & 28.0 & 64.7 & - \\
& SQuADv1.1 & 51.6 & 57.7 & 62.7 & 60.2 & 62.2 & 52.9 & 81.8 & 77.7 & 71.4 & 68.5 & 59.7 & 49.9 & 1.5 & 72.9 & 78.1 & 54.2 & 71.5 & 34.3 & 22.4 & 62.6 & - \\
& HellaSwag & 53.6 & 58.9 & 62.5 & 63.2 & 72.4 & 54.7 & 82.8 & 80.9 & 71.3 & 70.6 & 59.5 & 52.0 & 2.4 & 73.6 & 80.1 & 58.4 & 78.3 & 36.8 & 24.9 & 64.2 & - \\
& CCG & 54.6 & 53.5 & 60.6 & 62.8 & 69.1 & 41.6 & 80.7 & 78.1 & 65.4 & 68.1 & 55.1 & 51.6 & 1.3 & 68.7 & 79.8 & 61.9 & 68.8 & 37.9 & 19.8 & 61.6 & - \\
& Cosmos QA & 49.7 & 52.5 & 55.7 & 60.2 & 52.1 & 48.1 & 82.9 & 78.9 & 67.1 & 66.6 & 55.3 & 47.7 & 0.9 & 74.7 & 80.8 & 59.5 & 74.0 & 34.9 & 19.3 & 61.3 & - \\
& CSQA & 52.2 & 54.4 & 60.4 & 61.1 & 52.9 & 47.8 & \bf 83.4 & 80.7 & 68.5 & 69.0 & 57.9 & 50.1 & 1.4 & 73.6 & 81.5 & 63.2 & 74.0 & 43.6 & 19.3 & 62.9 & - \\
\bottomrule
\end{tabular}
}
\caption{Full NER Results}
\label{tab:full_panx}
\end{table*}

\begin{table*}[t!]
\resizebox{\textwidth}{!}{\small
\begin{tabular}{llcccccccccccc}
\toprule
&& ar & de & el & en & es & hi & ru & th & tr & vi & zh & Avg \\
\midrule
& XLM-R & 72.5 / 53.4 & 77.7 / 61.2 & 77.6 / 59.2 & 86.3 / 74.2 & 80.0 / 61.0 & 73.7 / 57.5 & 77.7 / 59.8 & 72.8 / 62.3 & 72.6 / 54.8 & 77.6 / 58.0 & 68.7 / 58.2 & 76.1 / 60.0 \\
\midrule \multirow{9}{*}{\STAB{\rotatebox[origin=c]{90}{\textbf{Without MLM}}}}
& \anli & 72.9 / 55.0 & 77.2 / 60.7 & 75.8 / 58.3 & 84.9 / 73.1 & 78.4 / 59.5 & 73.1 / 56.9 & 76.8 / 59.9 & 73.0 / 63.3 & 72.1 / 55.0 & 78.0 / 57.6 & 68.3 / 59.0 & 75.5 / 59.8 \\
& MNLI & 70.7 / 53.2 & 77.4 / 60.2 & 76.8 / 59.1 & 84.2 / 72.6 & 80.3 / 62.5 & 72.2 / 55.9 & 77.8 / 61.3 & 72.9 / 63.5 & 71.9 / 56.3 & 78.1 / 59.7 & 68.0 / 60.0 & 75.5 / 60.4 \\
& QQP & 68.4 / 50.4 & 73.2 / 56.5 & 73.3 / 55.9 & 82.3 / 70.6 & 75.4 / 57.3 & 68.5 / 52.5 & 74.2 / 57.5 & 68.6 / 60.2 & 68.3 / 51.4 & 72.9 / 53.4 & 66.3 / 58.0 & 72.0 / 56.7 \\
& SQuADv2.0 & 73.8 / 56.0 & 79.5 / 62.0 & 78.6 / 60.6 & 86.7 / 75.5 & 81.5 / {\bf 63.6} & 72.7 / 56.2 & 79.2 / 61.8 & 71.0 / 56.8 & 75.0 / 59.1 & 78.6 / 58.9 & 68.8 / 57.6 & 76.9 / 60.7 \\
& SQuADv1.1 & {\bf 75.9} / {\bf 59.9} & {\bf 80.3} / {\bf 63.6} & {\bf 80.3} / {\bf 62.1} & {\bf 88.3} / {\bf 77.4} & {\bf 81.8} / 63.2 & {\bf 76.1} / {\bf 59.2} & {\bf 80.0} / {\bf 64.1} & {\bf 75.6} / {\bf 65.5} & {\bf 75.8} / {\bf 59.2} & {\bf 80.5} / {\bf 61.2} & 70.8 / {\bf 61.3} & {\bf 78.7} / {\bf 63.3} \\
& HellaSwag & 73.9 / 56.9 & 78.7 / 61.3 & 77.9 / 58.8 & 86.1 / 75.6 & 79.6 / 60.1 & 74.3 / 57.5 & 78.5 / 62.8 & 73.6 / 64.5 & 73.5 / 56.6 & 78.8 / 59.1 & 69.2 / 59.4 & 76.7 / 61.1 \\
& CCG & 71.5 / 54.2 & 76.3 / 58.5 & 75.9 / 58.2 & 84.2 / 72.3 & 79.0 / 60.1 & 72.3 / 54.9 & 76.7 / 60.0 & 71.2 / 60.9 & 71.7 / 55.3 & 76.4 / 56.9 & 67.9 / 58.2 & 74.8 / 59.0 \\
& Cosmos QA & 73.2 / 53.8 & 78.1 / 62.2 & 77.3 / 58.3 & 86.7 / 75.4 & 79.9 / 61.9 & 74.2 / 57.7 & 77.9 / 59.4 & 72.3 / 61.5 & 73.3 / 55.6 & 78.2 / 58.0 & 68.3 / 58.5 & 76.3 / 60.2 \\
& CSQA & 72.6 / 53.4 & 79.5 / 62.4 & 78.3 / 59.4 & 87.1 / 76.1 & 81.0 / 62.9 & 74.9 / 58.5 & 77.6 / 60.3 & 69.7 / 58.9 & 73.4 / 56.5 & 78.2 / 58.1 & 67.5 / 57.3 & 76.3 / 60.3 \\
& Multi-task & 73.2 / 56.4 & 79.1 / 61.8 & 78.3 / 60.0 & 85.5 / 74.2 & 81.1 / 62.9 & 74.0 / 56.5 & 77.7 / 61.7 & 71.6 / 61.8 & 73.7 / 57.6 & 78.8 / 59.1 & 68.1 / 57.0 & 76.5 / 60.8 \\
\midrule \multirow{8}{*}{\STAB{\rotatebox[origin=c]{90}{\textbf{With MLM}}}}
& \anli & 72.1 / 52.4 & 77.3 / 59.8 & 76.1 / 57.6 & 85.8 / 74.1 & 78.7 / 58.8 & 72.9 / 55.3 & 76.9 / 59.4 & 73.0 / 63.4 & 72.3 / 55.3 & 78.5 / 57.8 & 70.9 / 61.0 & 75.9 / 59.5 \\
& MNLI & 72.5 / 54.8 & 78.4 / 60.7 & 77.8 / 60.4 & 86.4 / 75.5 & 80.4 / 61.3 & 73.6 / 56.6 & 78.2 / 61.7 & 73.9 / 64.5 & 72.5 / 57.5 & 79.0 / 60.3 & 69.0 / 59.7 & 76.5 / 61.2 \\
& QQP & 72.8 / 55.3 & 78.8 / 61.6 & 76.9 / 58.8 & 85.9 / 74.4 & 79.8 / 61.2 & 73.9 / 56.3 & 78.1 / 61.3 & 72.0 / 61.0 & 73.4 / 57.7 & 78.2 / 59.0 & 67.6 / 57.2 & 76.1 / 60.4 \\
& SQuADv2.0 & 72.3 / 55.0 & 79.0 / 63.3 & 76.9 / 58.6 & 85.3 / 73.9 & 80.3 / 61.9 & 73.1 / 56.9 & 77.8 / 61.7 & 72.5 / 61.1 & 72.8 / 55.8 & 77.8 / 58.2 & 68.4 / 58.6 & 76.0 / 60.4 \\
& SQuADv1.1 & 73.3 / 56.1 & 79.0 / 62.9 & 78.8 / 60.5 & 86.6 / 75.5 & 80.7 / 62.4 & 74.6 / 57.2 & 79.2 / 62.8 & 71.2 / 58.9 & 73.8 / 56.3 & 79.4 / 60.6 & 69.3 / 59.6 & 76.9 / 61.2 \\
& HellaSwag & 73.3 / 56.2 & 77.4 / 59.7 & 78.0 / 58.7 & 85.1 / 73.6 & 79.8 / 61.2 & 74.7 / 57.6 & 77.9 / 61.0 & 72.7 / 61.8 & 73.2 / 57.6 & 77.8 / 58.8 & 67.7 / 58.3 & 76.1 / 60.4 \\
& CCG & 71.8 / 53.2 & 77.4 / 60.5 & 75.7 / 56.9 & 84.8 / 72.9 & 79.3 / 60.1 & 73.1 / 55.8 & 75.8 / 57.1 & 70.3 / 58.3 & 71.7 / 55.6 & 77.2 / 57.0 & 66.9 / 57.4 & 74.9 / 58.6 \\
& Cosmos QA & 72.5 / 53.9 & 77.2 / 61.2 & 76.9 / 59.1 & 85.1 / 72.9 & 79.2 / 60.6 & 73.4 / 57.5 & 76.4 / 57.7 & 72.0 / 61.7 & 72.1 / 55.1 & 77.4 / 57.6 & 68.6 / 59.0 & 75.5 / 59.6 \\
& CSQA & 73.0 / 54.0 & 77.6 / 60.7 & 77.4 / 58.7 & 86.2 / 74.5 & 80.3 / 61.1 & 73.1 / 57.3 & 77.8 / 59.9 & 71.4 / 59.6 & 72.1 / 55.0 & 77.9 / 58.7 & {\bf 71.2} / 60.7 & 76.2 / 60.0 \\
\bottomrule
\end{tabular}
}
\caption{Full XQuAD Results}
\label{tab:full_xquad}
\end{table*}

\begin{table*}[t!]
\resizebox{\textwidth}{!}{\small
\begin{tabular}{llcccccccc}
\toprule
&& ar & de & en & es & hi & vi & zh & Avg \\
\midrule
& XLM-R & 62.7 / 42.4 & 69.1 / 52.0 & 81.6 / 68.6 & 72.2 / 53.0 & 68.0 / 50.7 & 69.5 / 47.6 & 67.9 / 46.2 & 70.1 / 51.5 \\
\midrule \multirow{9}{*}{\STAB{\rotatebox[origin=c]{90}{\textbf{Without MLM}}}}
& \anli & 64.1 / 43.9 & 66.8 / 49.8 & 82.5 / 69.4 & 71.9 / 52.6 & 69.2 / 50.5 & 70.5 / 49.7 & 66.9 / 44.8 & 70.3 / 51.5 \\
& MNLI & 64.2 / 43.5 & 68.1 / 51.8 & 82.7 / 70.0 & 73.7 / 54.8 & 70.3 / 52.7 & 68.9 / 49.5 & 67.1 / 46.0 & 70.7 / 52.6 \\
& QQP & 60.5 / 39.7 & 62.4 / 45.5 & 79.0 / 66.0 & 70.7 / 51.6 & 62.9 / 45.4 & 67.0 / 47.6 & 63.5 / 41.1 & 66.6 / 48.1 \\
& SQuADv2.0 & 66.1 / 45.3 & 68.2 / 50.2 & 83.5 / {\bf 71.1} & 73.6 / 55.4 & 68.5 / 51.5 & 71.7 / {\bf 52.4} & 68.2 / 46.4 & 71.4 / 53.2 \\
& SQuADv1.1 & {\bf 67.4} / {\bf 46.4} & {\bf 69.6} / 52.9 & {\bf 84.1} / 70.8 & {\bf 75.3} / {\bf 56.8} & {\bf 72.5} / {\bf 54.8} & 70.9 / 51.7 & {\bf 69.4} / 47.0 & {\bf 72.8} / {\bf 54.4} \\
& HellaSwag & 64.2 / 43.1 & 68.8 / 52.3 & 83.5 / 70.9 & 73.0 / 53.6 & 69.2 / 51.7 & 69.8 / 48.7 & 68.5 / 46.2 & 71.0 / 52.4 \\
& CCG & 62.7 / 41.6 & 67.5 / 50.4 & 82.9 / 70.0 & 72.9 / 54.6 & 66.1 / 50.1 & 68.9 / 48.9 & 66.4 / 45.6 & 69.6 / 51.6 \\
& Cosmos QA & 63.8 / 43.9 & 68.2 / 50.4 & 82.2 / 69.0 & 72.9 / 54.2 & 69.4 / 51.7 & 70.8 / 50.1 & 66.6 / 44.4 & 70.6 / 52.0 \\
& CSQA & 64.0 / 43.9 & 68.8 / 52.0 & 83.4 / 70.6 & 75.2 / 55.0 & 69.1 / 51.5 & {\bf 72.6} / 52.1 & 69.2 / 46.6 & 71.8 / 53.1 \\
& Multi-task & 65.1 / 44.1 & 70.2 / 54.9 & 82.9 / 69.4 & 75.2 / 56.4 & 70.1 / 52.3 & 72.0 / 51.7 & 68.6 / 46.2 & 72.0 / 53.6 \\
\midrule \multirow{8}{*}{\STAB{\rotatebox[origin=c]{90}{\textbf{With MLM}}}}
& \anli & 62.7 / 41.8 & 68.5 / 51.4 & 82.1 / 69.0 & 73.6 / 54.2 & 66.7 / 48.7 & 69.5 / 49.3 & 66.2 / 44.2 & 69.9 / 51.2 \\
& MNLI & 62.9 / 41.0 & 69.2 / {\bf 53.5} & 82.6 / 69.4 & 74.3 / 54.4 & 68.0 / 50.7 & 70.5 / 50.5 & 68.0 / 45.8 & 70.8 / 52.2 \\
& QQP & 64.6 / 44.9 & 68.1 / 51.2 & 83.2 / 70.4 & 74.0 / 55.6 & 70.4 / 53.1 & 69.1 / 49.3 & 68.3 / 45.6 & 71.1 / 52.9 \\
& SQuADv2.0 & 64.7 / 43.9 & 66.6 / 51.0 & 82.1 / 69.6 & 73.1 / 55.2 & 70.2 / 53.1 & 69.0 / 51.1 & 68.6 / {\bf 47.2} & 70.6 / 53.0 \\
& SQuADv1.1 & 64.4 / 43.3 & 68.0 / 50.0 & 83.1 / 70.0 & 75.2 / 56.2 & 68.5 / 51.9 & 71.2 / 51.9 & 66.8 / 44.6 & 71.0 / 52.6 \\
& HellaSwag & 64.7 / 44.3 & 68.4 / 52.3 & 83.3 / 70.4 & 73.9 / 55.0 & 69.5 / 52.1 & 69.9 / 47.9 & 67.7 / 44.8 & 71.1 / 52.4 \\
& CCG & 60.4 / 41.4 & 66.5 / 50.8 & 81.8 / 68.6 & 72.8 / 54.2 & 66.2 / 48.7 & 67.7 / 46.2 & 64.5 / 44.6 & 68.6 / 50.7 \\
& Cosmos QA & 63.4 / 43.1 & 69.0 / 51.0 & 81.9 / 68.9 & 72.3 / 53.6 & 66.3 / 48.9 & 69.1 / 47.6 & 66.0 / 45.2 & 69.7 / 51.2 \\
& CSQA & 64.3 / 43.7 & 69.5 / 51.8 & 82.6 / 69.4 & 73.4 / 54.4 & 68.0 / 50.7 & 70.9 / 48.7 & 67.7 / 45.8 & 70.9 / 52.1 \\
\bottomrule
\end{tabular}
}
\caption{Full MLQA Results}
\label{tab:full_mlqa}
\end{table*}

\begin{table*}[t!]
\resizebox{\textwidth}{!}{\small
\begin{tabular}{llcccccccccc}
\toprule
&& ar & bn & en & fi & id & ko & ru & sw & te & Avg \\
\midrule
& XLM-R & 64.5 / 46.9 & 59.5 / 41.6 & 70.4 / 56.6 & 64.9 / 49.2 & 75.1 / 59.8 & 54.7 / 39.5 & 65.4 / 43.6 & 67.2 / 48.7 & 68.8 / 48.3 & 65.6 / 48.2 \\
\midrule \multirow{8}{*}{\STAB{\rotatebox[origin=c]{90}{\textbf{Without MLM}}}}
& \anli & 67.3 / 47.8 & 54.9 / 37.2 & 71.0 / 57.3 & 64.7 / 47.8 & 74.9 / 57.5 & 54.5 / 41.3 & 62.4 / 33.0 & 67.2 / 47.3 & 68.2 / 46.9 & 65.0 / 46.2 \\
& MNLI & 67.8 / 49.7 & 60.6 / 40.7 & 71.6 / 57.7 & 66.5 / 48.6 & 76.6 / 61.9 & 55.3 / 42.4 & 63.9 / 39.0 & 66.9 / 48.5 & 71.0 / 51.4 & 66.7 / 48.9 \\
& QQP & 63.2 / 44.4 & 43.8 / 26.5 & 64.4 / 52.7 & 56.3 / 39.9 & 71.6 / 57.0 & 47.5 / 32.6 & 57.4 / 38.2 & 54.5 / 36.5 & 45.5 / 26.2 & 56.0 / 39.3 \\
& SQuADv2.0 & 76.5 / 59.8 & 77.7 / 63.7 & 76.1 / 63.2 & 78.3 / 64.3 & 83.1 / 69.9 & 68.1 / 56.5 & 73.0 / 51.5 & 79.1 / 67.1 & 79.2 / 61.1 & 76.8 / 61.9 \\
& SQuADv1.1 & 76.1 / 60.0 & 75.6 / 61.9 & 77.6 / 66.6 & 76.0 / 61.3 & 82.5 / 68.3 & 63.7 / 51.4 & 71.1 / 44.7 & 76.5 / 63.5 & 79.0 / 61.6 & 75.3 / 59.9 \\
& HellaSwag & 69.9 / 49.4 & 60.6 / 42.5 & 72.2 / 59.1 & 63.0 / 44.1 & 76.7 / 60.4 & 54.7 / 39.1 & 61.4 / 33.0 & 66.3 / 48.3 & 70.6 / 47.8 & 66.1 / 47.1 \\
& CCG & 63.6 / 41.8 & 54.1 / 37.2 & 68.5 / 55.9 & 59.6 / 41.7 & 73.2 / 57.5 & 50.8 / 37.7 & 60.2 / 33.4 & 66.8 / 49.7 & 66.2 / 43.8 & 62.6 / 44.3 \\
& Cosmos QA & 71.7 / 51.9 & 65.9 / 48.7 & 73.3 / 61.6 & 66.7 / 50.9 & 78.5 / 63.4 & 52.6 / 36.6 & 66.2 / 44.1 & 68.0 / 51.3 & 74.5 / 54.7 & 68.6 / 51.5 \\
& CSQA & 70.9 / 52.1 & 67.8 / 49.6 & 74.6 / 60.9 & 69.6 / 52.6 & 77.0 / 60.2 & 60.8 / 46.4 & 63.6 / 36.0 & 70.8 / 53.5 & 73.3 / 54.7 & 69.8 / 51.8 \\
& Multi-task & 73.3 / 52.3 & 66.7 / 48.7 & 75.6 / 63.6 & 74.7 / 59.6 & 81.7 / 67.3 & 60.2 / 46.4 & 71.0 / 43.0 & 76.0 / 64.3 & 77.2 / 58.4 & 72.9 / 56.0 \\
\midrule \multirow{8}{*}{\STAB{\rotatebox[origin=c]{90}{\textbf{With MLM}}}}
& \anli & 67.1 / 48.9 & 59.5 / 42.5 & 72.2 / 58.9 & 67.2 / 51.4 & 76.8 / 60.7 & 54.9 / 42.0 & 62.4 / 35.3 & 70.3 / 52.1 & 70.4 / 53.1 & 66.8 / 49.4 \\
& MNLI & 67.3 / 49.7 & 60.0 / 41.6 & 71.2 / 59.3 & 66.8 / 50.4 & 78.1 / 62.1 & 56.4 / 42.0 & 62.2 / 33.9 & 68.5 / 50.7 & 70.0 / 48.4 & 66.7 / 48.7 \\
& QQP & 67.8 / 49.0 & 55.7 / 37.2 & 69.8 / 56.1 & 64.1 / 47.1 & 74.2 / 58.6 & 49.0 / 34.4 & 60.0 / 34.5 & 64.5 / 45.7 & 70.1 / 45.6 & 63.9 / 45.3 \\
& SQuADv2.0 & 76.9 / 60.5 & 70.1 / 54.9 & 76.6 / 64.5 & 74.4 / 59.6 & 83.4 / 69.7 & 61.6 / 48.6 & 71.3 / 45.2 & 74.0 / 61.5 & 76.7 / 59.3 & 73.9 / 58.2 \\
& SQuADv1.1 & 77.0 / 59.3 & 68.5 / 51.3 & 75.4 / 64.3 & 77.2 / 63.4 & 83.3 / 71.0 & 63.7 / 51.8 & 71.7 / 47.9 & 73.1 / 56.5 & 76.4 / 59.0 & 74.0 / 58.3 \\
& HellaSwag & 68.8 / 50.4 & 62.6 / 47.8 & 70.9 / 56.8 & 64.0 / 48.6 & 77.4 / 61.8 & 54.6 / 40.9 & 61.2 / 31.7 & 68.2 / 49.5 & 71.4 / 50.5 & 66.6 / 48.7 \\
& CCG & 68.1 / 49.1 & 57.5 / 39.8 & 69.0 / 55.9 & 65.9 / 48.6 & 76.5 / 61.9 & 55.0 / 39.9 & 61.6 / 31.9 & 67.5 / 49.3 & 56.3 / 30.3 & 64.2 / 45.2 \\
& Cosmos QA & 66.6 / 46.6 & 56.8 / 37.2 & 71.5 / 58.0 & 64.2 / 45.0 & 75.0 / 57.0 & 56.3 / 41.3 & 63.6 / 39.0 & 69.0 / 51.1 & 63.6 / 46.3 & 65.2 / 46.8 \\
& CSQA & 68.8 / 50.4 & 60.2 / 43.4 & 71.3 / 59.1 & 67.6 / 50.5 & 76.9 / 59.8 & 54.0 / 41.3 & 63.5 / 38.1 & 69.5 / 52.9 & 72.8 / 54.1 & 67.2 / 49.9 \\
\bottomrule
\end{tabular}
}
\caption{Full TyDiQA Results}
\label{tab:full_tydiqa}
\end{table*}

\begin{table*}[t!]
\centering\small
\begin{tabular}{llccccc}
\toprule
&& de & fr & ru & zh & Avg \\
\midrule
& XLM-R & 77.7 & 62.7 & 79.2 & 66.5 & 71.5 \\
\midrule \multirow{9}{*}{\STAB{\rotatebox[origin=c]{90}{\textbf{Without MLM}}}}
& \anli & 94.6 & 89.8 & 93.5 & 88.6 & 91.6 \\
& MNLI & 94.2 & 90.2 & 93.5 & \bf 89.9 & 92.0 \\
& QQP & 94.2 & 91.0 & 93.3 & 88.5 & 91.8 \\
& SQuADv2.0 & 94.0 & 89.8 & 93.0 & \bf 89.9 & 91.7 \\
& SQuADv1.1 & 94.2 & 90.5 & 93.1 & 87.0 & 91.2 \\
& HellaSwag & 94.6 & \bf 91.9 & \bf 93.9 & 88.9 & \bf 92.3 \\
& CCG & 88.3 & 82.9 & 86.6 & 78.0 & 83.9 \\
& Cosmos QA & 94.1 & 90.2 & 93.2 & 88.6 & 91.5 \\
& CSQA & \bf 95.1 & 90.6 & 93.5 & 89.1 & 92.1 \\
& Multi-task & 94.3 & 90.4 & 93.4 & 87.0 & 91.3 \\
\midrule \multirow{8}{*}{\STAB{\rotatebox[origin=c]{90}{\textbf{With MLM}}}}
& \anli & 93.4 & 88.0 & 92.9 & 86.5 & 90.2 \\
& MNLI & 92.7 & 89.0 & 93.2 & 86.1 & 90.3 \\
& QQP & 90.8 & 86.9 & 90.6 & 83.6 & 88.0 \\
& SQuADv2.0 & 92.8 & 87.0 & 91.4 & 85.8 & 89.2 \\
& SQuADv1.1 & 92.9 & 89.5 & 92.7 & 85.3 & 90.1 \\
& HellaSwag & 92.6 & 87.5 & 91.4 & 86.6 & 89.5 \\
& CCG & 87.6 & 78.5 & 87.6 & 75.7 & 82.4 \\
& Cosmos QA & 91.8 & 86.9 & 91.7 & 88.4 & 89.7 \\
& CSQA & 86.1 & 80.8 & 87.9 & 81.6 & 84.1 \\

\bottomrule
\end{tabular}
\caption{Full BUCC Results}
\label{tab:full_bucc2018}
\end{table*}

\begin{table*}[t!]
\resizebox{\textwidth}{!}{\small
\begin{tabular}{llccccccccccccccccccc}
\toprule
&& af & ar & bg & bn & de & el & es & et & eu & fa & fi & fr & he & hi & hu & id & it & ja & jv \\
\midrule
& XLM-R & 30.5 & 20.4 & 39.0 & 13.3 & 63.9 & 18.9 & 48.0 & 25.8 & 19.9 & 42.0 & 41.5 & 48.1 & 28.0 & 38.3 & 42.5 & 47.0 & 42.3 & 41.8 & 10.2 \\
\midrule \multirow{9}{*}{\STAB{\rotatebox[origin=c]{90}{\textbf{Without MLM}}}}
& \anli & 78.8 & 74.0 & 88.0 & 72.3 & 97.4 & 82.4 & 91.2 & 70.9 & 53.3 & 91.5 & 88.6 & 89.8 & 82.1 & 92.8 & 86.2 & \bf 92.1 & 82.6 & 88.7 & 31.7 \\
& MNLI & 79.6 & 70.7 & 84.8 & 71.2 & 96.6 & 82.5 & 93.1 & 74.3 & 59.2 & 90.0 & 89.0 & 89.6 & 81.8 & 91.7 & 86.0 & 91.7 & 86.3 & 89.5 & 30.7 \\
& QQP & \bf 80.4 & 74.9 & 87.3 & 74.3 & 96.5 & 84.1 & \bf 93.8 & 74.7 & 60.2 & 91.0 & 90.3 & 89.9 & \bf 86.0 & 93.3 & 88.4 & \bf 92.1 & 86.3 & 89.9 & 35.6 \\
& SQuADv2.0 & 73.7 & 67.7 & 84.2 & 63.2 & 96.0 & 74.3 & 89.2 & 70.5 & 54.0 & 87.9 & 85.5 & 87.1 & 77.1 & 88.0 & 83.5 & 89.5 & 80.2 & 86.4 & 32.2 \\
& SQuADv1.1 & 76.9 & 68.9 & 85.7 & 65.7 & 96.4 & 76.3 & 89.5 & 76.9 & 58.4 & 88.0 & 88.5 & 88.5 & 77.3 & 89.9 & 84.0 & 90.4 & 83.0 & 88.7 & 30.2 \\
& HellaSwag & 78.9 & \bf 75.4 & \bf 89.9 & \bf 75.4 & \bf 97.7 & \bf 84.8 & 93.1 & \bf 79.8 & 64.8 & \bf 91.8 & \bf 92.0 & \bf 92.2 & 84.9 & \bf 93.4 & \bf 89.5 & \bf 92.1 & \bf 86.7 & \bf 91.6 & \bf 37.1 \\
& CCG & 71.9 & 59.1 & 82.1 & 62.5 & 95.5 & 74.4 & 87.0 & 67.3 & 49.0 & 84.7 & 82.6 & 84.4 & 77.2 & 85.4 & 80.7 & 87.2 & 79.1 & 78.7 & 24.9 \\
& Cosmos QA & 78.6 & 70.6 & 86.6 & 71.0 & 96.4 & 80.5 & 91.8 & 77.6 & 60.7 & 89.8 & 91.3 & 89.4 & 83.0 & 91.5 & 87.7 & 91.4 & 83.7 & 88.2 & 37.1 \\
& CSQA & 79.5 & 74.5 & 87.7 & 74.0 & 96.9 & 83.6 & 92.9 & 79.1 & 65.8 & 90.0 & \bf 92.0 & 90.7 & 83.1 & 92.2 & 88.4 & 91.8 & 85.4 & 88.9 & 33.7 \\
& Multi-task & 81.2 & 71.9 & 88.0 & 73.6 & 97.1 & 82.9 & 92.6 & 73.1 & 58.6 & 90.4 & 89.6 & 89.6 & 84.1 & 92.6 & 87.2 & 92.6 & 83.9 & 91.0 & 34.1 \\
\midrule
\midrule \multirow{8}{*}{\STAB{\rotatebox[origin=c]{90}{\textbf{With MLM}}}}
& \anli & 78.6 & 65.2 & 86.6 & 67.8 & 97.0 & 78.2 & 90.2 & 79.1 & 59.3 & 89.3 & 89.1 & 90.4 & 78.7 & 89.3 & 86.5 & 91.0 & 84.6 & 87.0 & 26.3 \\
& MNLI & 77.3 & 65.2 & 83.8 & 64.9 & 97.2 & 76.1 & 92.1 & 77.7 & 57.3 & 88.1 & 88.8 & 87.5 & 81.0 & 89.0 & 87.1 & 90.5 & 82.6 & 85.6 & 27.3 \\
& QQP & 74.4 & 61.3 & 83.7 & 64.6 & 96.2 & 75.7 & 88.1 & 76.7 & 59.4 & 86.3 & 87.0 & 86.9 & 76.6 & 85.9 & 84.2 & 89.8 & 79.8 & 84.0 & 28.8 \\
& SQuADv2.0 & 70.8 & 57.6 & 80.9 & 52.7 & 96.6 & 63.4 & 84.5 & 71.5 & 47.4 & 85.4 & 86.9 & 85.1 & 71.9 & 85.2 & 83.9 & 90.4 & 78.1 & 83.2 & 16.1 \\
& SQuADv1.1 & 79.2 & 67.7 & 86.5 & 71.4 & 96.7 & 80.4 & 91.6 & 83.1 & \bf 66.3 & 90.8 & 91.1 & 89.8 & 77.5 & 92.3 & 87.4 & 91.8 & 84.6 & 87.4 & 26.3 \\
& HellaSwag & 57.1 & 45.2 & 69.4 & 40.4 & 89.7 & 57.8 & 73.4 & 64.0 & 42.2 & 77.1 & 76.4 & 76.5 & 62.6 & 75.1 & 76.2 & 82.5 & 69.7 & 77.5 & 22.0 \\
& CCG & 71.9 & 52.3 & 80.4 & 51.0 & 95.0 & 72.6 & 86.0 & 73.5 & 51.0 & 83.3 & 84.1 & 81.8 & 71.3 & 79.1 & 81.6 & 87.2 & 78.7 & 76.2 & 12.7 \\
& Cosmos QA & 69.7 & 63.7 & 84.0 & 58.8 & 95.1 & 74.2 & 84.6 & 76.5 & 58.6 & 85.7 & 85.2 & 84.5 & 76.2 & 87.1 & 84.7 & 88.5 & 81.4 & 85.5 & 24.9 \\
& CSQA & 54.3 & 45.3 & 63.6 & 33.5 & 87.0 & 50.5 & 70.0 & 58.8 & 35.7 & 74.1 & 71.0 & 70.7 & 58.2 & 70.2 & 72.5 & 80.4 & 64.2 & 75.5 & 16.6 \\
\midrule
\midrule
&& ka & kk & ko & ml & mr & nl & pt & ru & sw & ta & te & th & tl & tr & ur & vi & zh & Avg & \\
\midrule
& XLM-R & 11.8 & 17.4 & 35.5 & 19.4 & 15.2 & 52.6 & 47.2 & 42.1 & 7.9 & 9.1 & 19.7 & 27.4 & 10.3 & 37.8 & 22.5 & 38.3 & 41.2 & 31.0 & - \\
\midrule \multirow{9}{*}{\STAB{\rotatebox[origin=c]{90}{\textbf{Without MLM}}}}
& \anli & 76.9 & 67.3 & 84.6 & 90.8 & 80.5 & \bf 93.6 & 91.0 & 90.5 & 30.8 & 76.5 & 85.5 & \bf 91.2 & 59.9 & 87.9 & 79.7 & 94.6 & \bf 93.0 & 80.8 & - \\
& MNLI & 77.9 & 67.7 & 84.3 & 89.8 & 80.4 & 92.5 & 91.3 & 89.2 & 32.8 & 70.0 & 78.2 & 86.7 & 60.9 & 88.8 & 74.5 & 92.5 & 91.2 & 80.2 & - \\
& QQP & 78.7 & 69.4 & 86.4 & \bf 92.9 & \bf 82.9 & 93.3 & \bf 92.5 & 91.6 & 35.1 & \bf 81.4 & \bf 90.6 & 90.0 & 64.6 & \bf 91.4 & 81.7 & 95.0 & 92.3 & 82.7 & - \\
& SQuADv2.0 & 67.0 & 63.0 & 80.8 & 82.8 & 71.6 & 89.7 & 90.4 & 86.9 & 27.7 & 60.9 & 74.4 & 80.7 & 54.2 & 85.9 & 70.6 & 92.5 & 89.3 & 76.1 & - \\
& SQuADv1.1 & 70.9 & 63.7 & 83.3 & 87.3 & 74.7 & 91.7 & 90.2 & 89.1 & 31.5 & 60.6 & 77.8 & 82.3 & 59.3 & 88.3 & 68.3 & 92.8 & 90.8 & 77.9 & - \\
& HellaSwag & \bf 80.8 & \bf 72.0 & \bf 86.5 & 92.1 & 81.1 & 93.2 & 91.9 & \bf 92.0 & 35.1 & 79.2 & 87.2 & 89.6 & 64.5 & 90.6 & \bf 82.4 & \bf 95.1 & 92.6 & \bf 83.3 & - \\
& CCG & 65.1 & 56.9 & 76.8 & 82.5 & 70.3 & 88.9 & 88.8 & 84.5 & 24.9 & 60.3 & 65.4 & 72.8 & 53.3 & 82.6 & 64.7 & 89.7 & 84.8 & 72.9 & - \\
& Cosmos QA & 75.7 & 69.9 & 83.6 & 90.1 & 78.7 & 92.0 & 91.3 & 89.7 & 34.1 & 72.3 & 84.6 & 89.1 & 59.7 & 89.6 & 79.8 & 93.3 & 90.9 & 80.9 & - \\
& CSQA & \bf 80.8 & 70.3 & 85.5 & 91.7 & 82.7 & 93.3 & 91.4 & 90.4 & \bf 35.9 & 73.3 & 84.6 & 89.4 & \bf 65.4 & 90.2 & 77.1 & 94.8 & 92.9 & 82.2 & - \\
& Multi-task & 78.7 & 68.2 & 85.0 & 91.4 & 80.4 & 92.1 & 92.0 & 90.2 & 34.4 & 68.7 & 83.8 & 89.1 & 62.3 & 88.9 & 77.6 & 95.0 & 92.8 & 81.2 & - \\
\midrule
\midrule \multirow{9}{*}{\STAB{\rotatebox[origin=c]{90}{\textbf{With MLM}}}}
& \anli & 70.6 & 64.7 & 83.6 & 88.9 & 75.6 & 92.0 & 91.0 & 88.1 & 29.0 & 70.0 & 76.9 & 84.7 & 51.6 & 88.0 & 71.7 & 93.6 & 91.6 & 78.5 & - \\
& MNLI & 67.7 & 63.3 & 81.8 & 84.3 & 75.0 & 90.8 & 90.5 & 87.8 & 29.7 & 62.2 & 73.5 & 85.2 & 53.4 & 87.6 & 71.2 & 93.3 & 88.5 & 77.4 & - \\
& QQP & 66.0 & 64.2 & 80.2 & 82.0 & 70.6 & 89.4 & 89.8 & 86.7 & 30.5 & 60.9 & 76.1 & 83.6 & 52.3 & 84.9 & 72.7 & 90.5 & 88.0 & 76.0 & - \\
& SQuADv2.0 & 53.8 & 54.8 & 77.5 & 72.5 & 61.5 & 90.0 & 87.0 & 87.2 & 20.3 & 41.7 & 51.7 & 80.5 & 38.0 & 81.8 & 63.3 & 90.6 & 89.1 & 70.4 & - \\
& SQuADv1.1 & 73.2 & 66.8 & 83.9 & 89.8 & 78.9 & 93.0 & 90.4 & 89.7 & 33.8 & 76.2 & 85.0 & 90.0 & 54.5 & 90.0 & 78.6 & 93.6 & 90.9 & 80.6 & - \\
& HellaSwag & 38.5 & 43.1 & 70.5 & 63.2 & 39.7 & 79.1 & 78.4 & 80.0 & 19.2 & 30.9 & 55.6 & 66.6 & 33.1 & 71.5 & 49.8 & 80.4 & 77.7 & 61.4 & - \\
& CCG & 58.3 & 51.3 & 74.6 & 76.3 & 58.4 & 89.0 & 86.9 & 82.9 & 23.3 & 46.9 & 60.3 & 72.6 & 40.9 & 82.5 & 55.8 & 87.9 & 80.3 & 69.4 & - \\
& Cosmos QA & 63.3 & 56.0 & 80.7 & 79.0 & 63.1 & 89.4 & 87.2 & 86.1 & 26.2 & 55.7 & 71.8 & 80.5 & 44.6 & 83.0 & 63.7 & 91.0 & 85.1 & 73.8 & - \\
& CSQA & 33.4 & 36.2 & 65.9 & 47.0 & 30.9 & 76.6 & 74.7 & 75.5 & 19.0 & 28.3 & 49.6 & 64.1 & 26.0 & 64.1 & 53.0 & 78.4 & 75.1 & 56.9 & - \\
\bottomrule
\end{tabular}
}
\caption{Full Tatoeba Results}
\label{tab:full_tatoeba}
\end{table*}

\begin{table*}[t!]
\centering \small
\begin{tabular}{lccc}
\toprule
& MNLI & QQP & HellaSwag \\
\midrule
\texttt{en} & 87.1 & 88.0 & 71.6 \\
Translated to \texttt{de} & 82.2 & 84.6& 55.1 \\
Translated to \texttt{ru} & 70.1 & 83.8 & 27.4 \\
Translated to \texttt{sw} & 70.8 & 79.3 & 25.1 \\
\bottomrule
\end{tabular}
\caption{Intermediate task performance on trained and evaluated on translated data. We report the median result for English (original) task data.}
\label{tab:translated_intermediate_performance}
\end{table*}

\end{document}